\documentclass[letterpaper, 10 pt, conference]{ieeeconf}  % Comment this line out
\IEEEoverridecommandlockouts                              % This command is only
\usepackage[utf8]{inputenc}
\usepackage[T1]{fontenc}
\usepackage{graphicx}
\usepackage{float} 
\usepackage{subfigure}
\usepackage{ragged2e}
\usepackage{booktabs}
\usepackage{bbding}
\usepackage{multirow}
\usepackage{makecell}
\usepackage[ruled,vlined]{algorithm2e}
\usepackage{amsmath,amssymb,amsfonts}
\usepackage[]{caption2} %新增调用的宏包
\usepackage[table,xcdraw]{xcolor}
\usepackage{cite}
\usepackage{mathptmx} % assumes new font selection scheme installed
\usepackage{mathptmx} % assumes new font selection scheme installed
\usepackage[colorlinks,linkcolor=blue]{hyperref}

\usepackage{fancyhdr}
\usepackage{dsfont}
\usepackage{array,color}
\usepackage{bm}
\usepackage{algpseudocode}
\usepackage{bbm}

\title{\LARGE \bf
Communication Resources Constrained Hierarchical Federated Learning for End-to-End Autonomous Driving
}

\author{Wei-Bin Kou$^{1,3,4}$, Shuai Wang$^{2,3,*}$, Guangxu Zhu$^{4}$, Bin Luo$^{5}$,\\Yingxian Chen$^{1}$, Derrick Wing Kwan Ng$^{6}$, and Yik-Chung Wu$^{1,*}$
\thanks{This work has been accepted by 2023 IEEE/RSJ International Conference on Intelligent Robots and Systems (IROS). This work was supported in part by the Science and Technology Development Fund of Macao S.A.R (FDCT) (No. 0081/2022/A2), in part by Open Research Fund from Guangdong Laboratory of Artificial Intelligence and Digital Economy (SZ) (No. GML-KF-22-17), in part by the SIAT Direct Drive Tech Cooperation Project, in part by National Natural Science Foundation of China (No. 62001310), in part by Guangdong Basic and Applied Basic Research Foundation (No. 2022A1515010109) and in part by the Internal Project Fund from Shenzhen Research Institute of Big Data (No. J00120230001).}
\thanks{$^{*}$Corresponding author: Shuai Wang (s.wang@siat.ac.cn) and Yik-Chung Wu (ycwu@eee.hku.hk).}% <-this % stops a space
\thanks{$^{1}$Department of Electrical and Electronic Engineering, The University of Hong Kong, Hong Kong 999077, China.}%
\thanks{$^{2}$Guangdong Laboratory of Artificial Intelligence and Digital Economy (SZ), Shenzhen, China.}%
\thanks{$^{3}$Shenzhen Institute of Advanced Technology, Chinese Academy of Sciences, Shenzhen, China.}%
\thanks{$^{4}$Shenzhen Research Institute of Big Data, Shenzhen, China.}%
\thanks{$^{5}$Damo Institute, Alibaba, China.}%
\thanks{$^{6}$School of Electrical Engineering and Telecommunications, the University of New South Wales, Australia}
}

\begin{document}

\maketitle
\thispagestyle{empty}
\pagestyle{empty}

%%%%%%%%%%%%%%%%%%%%%%%%%%%%%%%%%%%%%%%%%%%%%%%%%%%%%%%%%%%%%%%%%%%%%%%%%%%%%%%%
\begin{abstract}
While federated learning (FL) improves the generalization of end-to-end autonomous driving by model aggregation, the conventional single-hop FL (SFL) suffers from slow convergence rate due to long-range communications among vehicles and cloud server. 
Hierarchical federated learning (HFL) overcomes such drawbacks via introduction of mid-point edge servers. 
However, the orchestration between constrained communication resources and HFL performance becomes an urgent problem. This paper proposes an optimization-based Communication Resource Constrained Hierarchical Federated Learning (CRCHFL) framework to minimize the generalization error of the autonomous driving model using hybrid data and model aggregation. 
The effectiveness of the proposed CRCHFL is evaluated in the Car Learning to Act (CARLA) simulation platform.
Results show that the proposed CRCHFL both accelerates the convergence rate and enhances the generalization of federated learning autonomous driving model. Moreover, under the same communication resource budget, it outperforms the HFL by 10.33\% and the SFL by 12.44\%.
\end{abstract}
\vspace{-0.15cm}
%%%%%%%%%%%%%%%%%%%%%%%%%%%%%%%%%%%%%%%%%%%%%%%%%%%%%%%%%%%%%%%%%%%%%%%%%%%%%%%%
\section{INTRODUCTION}
\vspace{-0.15cm}
Vision-based autonomous driving vehicles, such as Tesla, are increasingly prevalent in our life, but the generalization needs to be continuously enhanced because the embedded autonomous driving model cannot generalize to all scenarios. 
Federated learning (FL) is an emerging paradigm to overcome the domain shifting issue to enhance generalization\cite{DBLP:journals/corr/abs-2111-10487} in end-to-end autonomous driving (FLEAD).
Despite the fact that many efforts have been invested in developing FLEAD techniques \cite{DBLP:journals/corr/abs-1910-06001,DBLP:journals/corr/abs-1901-06455,9834117, https://doi.org/10.48550/arxiv.2206.01748,DBLP:journals/corr/abs-2103-03786, DBLP:journals/corr/abs-2103-11879, DBLP:journals/corr/abs-2110-05754, DBLP:journals/corr/abs-2010-08303, Savazzi2021OpportunitiesOF}, a number of technical challenges still need to be properly handled: 
I) \textbf{Limited communication resources}. Current FLEAD designs aim to maximize the driving performance without any communication constraints, thus ignoring the interdependency between driving performance and practical communication conditions. II) \textbf{Tradeoff between learning performance and communication cost}. How can we effectively allocate the network resources to different vehicles, edge servers and cloud server to maximize the FL model performance while satisfying the stringent communication constraints? 

To tackle these challenges, currently there are three methods mainly to compress model parameters or gradient size: quantization-based, sparsification-based, and distillation-based strategies\cite{https://doi.org/10.48550/arxiv.2208.01200}. Specifically, quantization-based methods \cite{Gray2022Quantization} try to quantize continuous model parameters or gradient into a discrete set to reduce the bits. For sparsification, it converts parameters or gradient to a sparse one or zero according to importance of corresponding elements. Top-k and rand-k sparsification are two widely utilized approaches\cite{DBLP:journals/corr/abs-2009-09271}. Distillation\cite{https://doi.org/10.48550/arxiv.1503.02531} is proposed to transfer a large model (teacher) to a small model (student) without obvious performance loss. In summary, these mainstream methods to mitigate the FL communication overheads is by reducing the model parameters or gradient size, leading to some performance loss\cite{DBLP:journals/corr/abs-2202-02812} compared to the case without any compression. 

In this paper, we propose to maximize the hierarchical federated learning (HFL) model performance under constrained communication resources from a completely different perspective: optimization-based communication resource scheduling under limited budgets. To the best of our knowledge, this is the first work to integrate constrained throughput scheduling and HFL in FLEAD. To begin with, we propose an optimization algorithm to distinguish the priorities of different HFL stages and perform resource scheduling with consideration of both data and model transfer. This algorithm supports both sample size and FL round planning. In addition, based on this optimization algorithm, we elaborate communication resources constrained HFL framework (CRCHFL) to enhance convergence and generalization by leveraging both data and model parameter aggregation. Specifically, it consists of: I). \textbf{cloud pretraining stage} to collect data from vehicles to cloud server for centralized pretraining and then to release pretrained model to all edge servers and vehicles to accelerate FL convergence; II). \textbf{edge federated learning stage} to aggregate model parameters of all associated vehicles; and III). \textbf{cloud federated learning stage} to aggregate model parameters of all edge servers. In particular, our main contributions are:
\begin{itemize}
\item[(1)] We propose CRCHFL framework to maximize the model performance under constrained communication resources.

\item[(2)] Evaluation and analysis of our proposed approach on simulation dataset.

\item[(3)] Implementing the proposed CRCHFL scheme located at  (\href{https://github.com/WeibinKOU/CRCHFL}{https://github.com/WeibinKOU/CRCHFL.git}) based on the high-fidelity CARLA\cite{dosovitskiy2017carla} simulator.
\end{itemize}

\section{Related Work}
Current autonomous driving (AD) systems can be divided into two categories: modular-based \cite{7535474, jiang2019multi} and learning-based \cite{DBLP:journals/corr/abs-1810-02890, DBLP:journals/corr/abs-1905-00229, doi:10.1177/0278364919880273,9013081,9165167, 9834117, https://doi.org/10.48550/arxiv.2206.01748,DBLP:journals/corr/abs-2103-03786, DBLP:journals/corr/abs-2103-11879, DBLP:journals/corr/abs-2110-05754, DBLP:journals/corr/abs-2010-08303}. Modular-based methods result in error propagation and inaccuracies in both problem modelling and solving stages.
End-to-end learning-based approaches can address this error propagation issue. These methods generally map the onboard sensor data, such as LiDAR point clouds and camera images, into driving actions, such as throttle, brake and steer directly. However, such learning-based methods would have generalization issue intrinsically and work only in limited scenarios. 

FL is an emerging paradigm to improve the generalization of end-to-end learning-based methods via model parameter aggregation or data collaboration\cite{DBLP:journals/corr/abs-2010-08303}. In the context of EAD, FL leverages vehicular networks to integrate knowledge from different vehicles located in various scenarios.
As such, when an EAD system enters a new scenario, it can convey the knowledge of new samples or corner cases to other vehicles and remote servers while preserving data privacy \cite{9013081, 9834117, https://doi.org/10.48550/arxiv.2206.01748,DBLP:journals/corr/abs-2103-03786, DBLP:journals/corr/abs-2103-11879, DBLP:journals/corr/abs-2110-05754, DBLP:journals/corr/abs-2010-08303}.
For example, in \cite{9013081}, a cloud federated robotic system is proposed to enhance the behaviour cloning method, generating accurate control commands based on RGB images, depth images and semantic segmentation images. 

Nonetheless, such single-hop federated learning (SFL) converges slow owing to its long communication latency\cite{DBLP:journals/corr/abs-2010-11612}. HFL can accelerate the training procedure since edge servers are closer to vehicles and more communication-efficient than cloud server.
On the other hand, for HFL, there exists frequent communication flows among different nodes, such as vehicles, edge servers, and cloud server. 
Such communications among nodes in \cite{9013081, 9834117,DBLP:journals/corr/abs-2103-03786, DBLP:journals/corr/abs-2103-11879, DBLP:journals/corr/abs-2110-05754, DBLP:journals/corr/abs-2010-08303} are assumed to be perfect, which does not hold for practical EAD systems with limited resources.
The mainstream approaches address the constrained communication resource problem of HFL in practical situations by compressing the model parameter or gradient size. There are now three main approaches:  quantization-based method (e.g., \cite{DBLP:journals/corr/Alistarh0TV16}),  sparsification-based method (e.g., \cite{DBLP:journals/corr/abs-2108-00951}), and distillation-based method (e.g., \cite{DBLP:journals/corr/abs-1910-03581}). It is worth noting that most of these methods come at the expense of a certain amount of performance. 
Therefore, it becomes imperative to develop associated methods to allocate the limited network resources for HFL to minimize the generalization error of deep neural networks (DNNs) without any performance loss.
This inspires us to design a CRCHFL framework, which fuses both data and model parameters and optimizes limited communication resources, to maximize the FLEAD convergence and generalization performance.

\begin{figure}[!t]
\vspace{-0.3cm}
\centering 
\includegraphics[width=0.48\textwidth, height=0.24\textwidth]{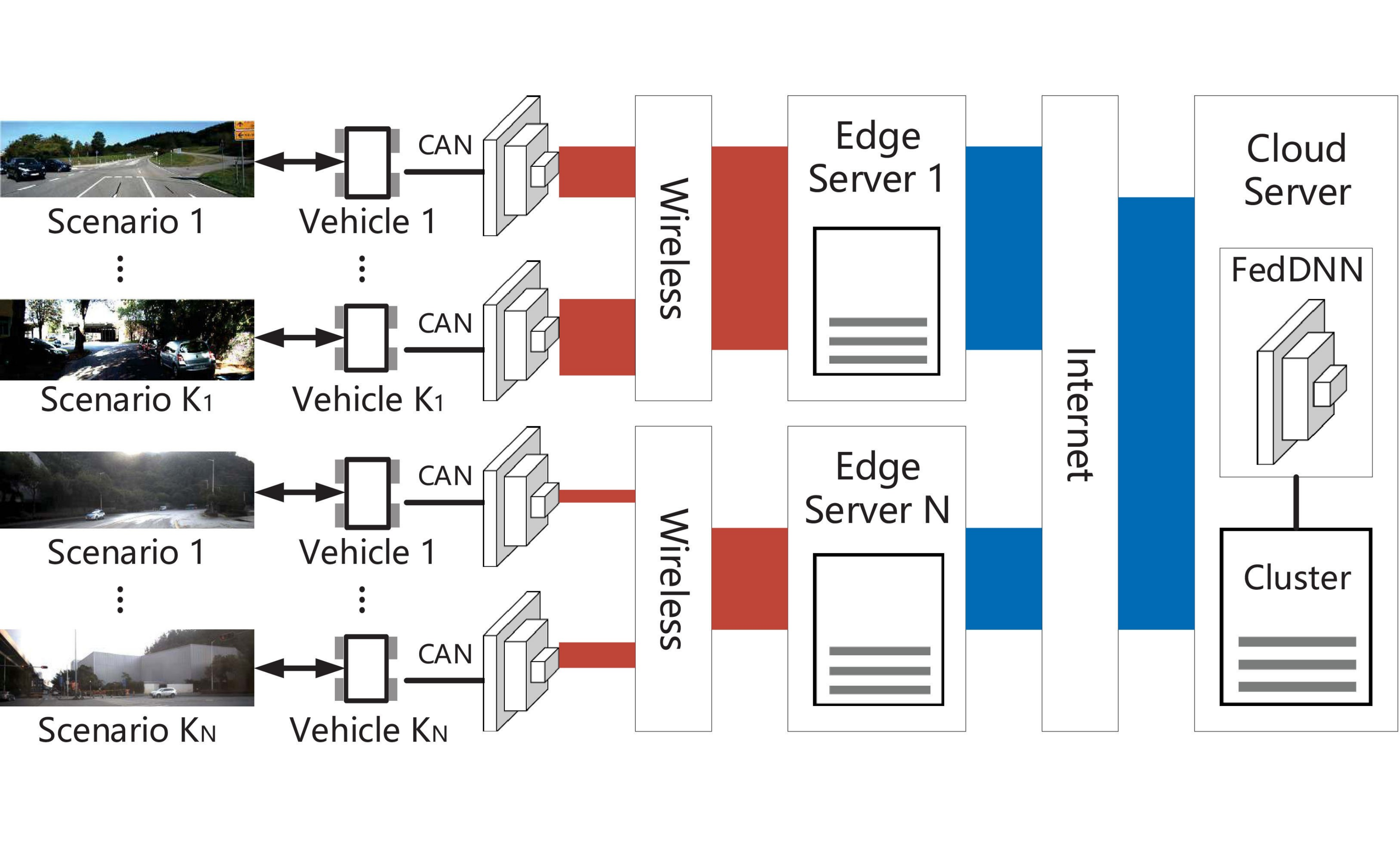}
\vspace{-0.9cm}
\caption{Illustration of a cloud-edge-vehicle system. Red bars represent wireless flows. Blue bars represent wireline flows. The size of each bar represents the communication throughput of its associated link.}
\vspace{-0.7cm}
\label{Fig.main}
\end{figure}

\begin{figure*}[!t]
\centering
\includegraphics[width=180mm, height=65mm]{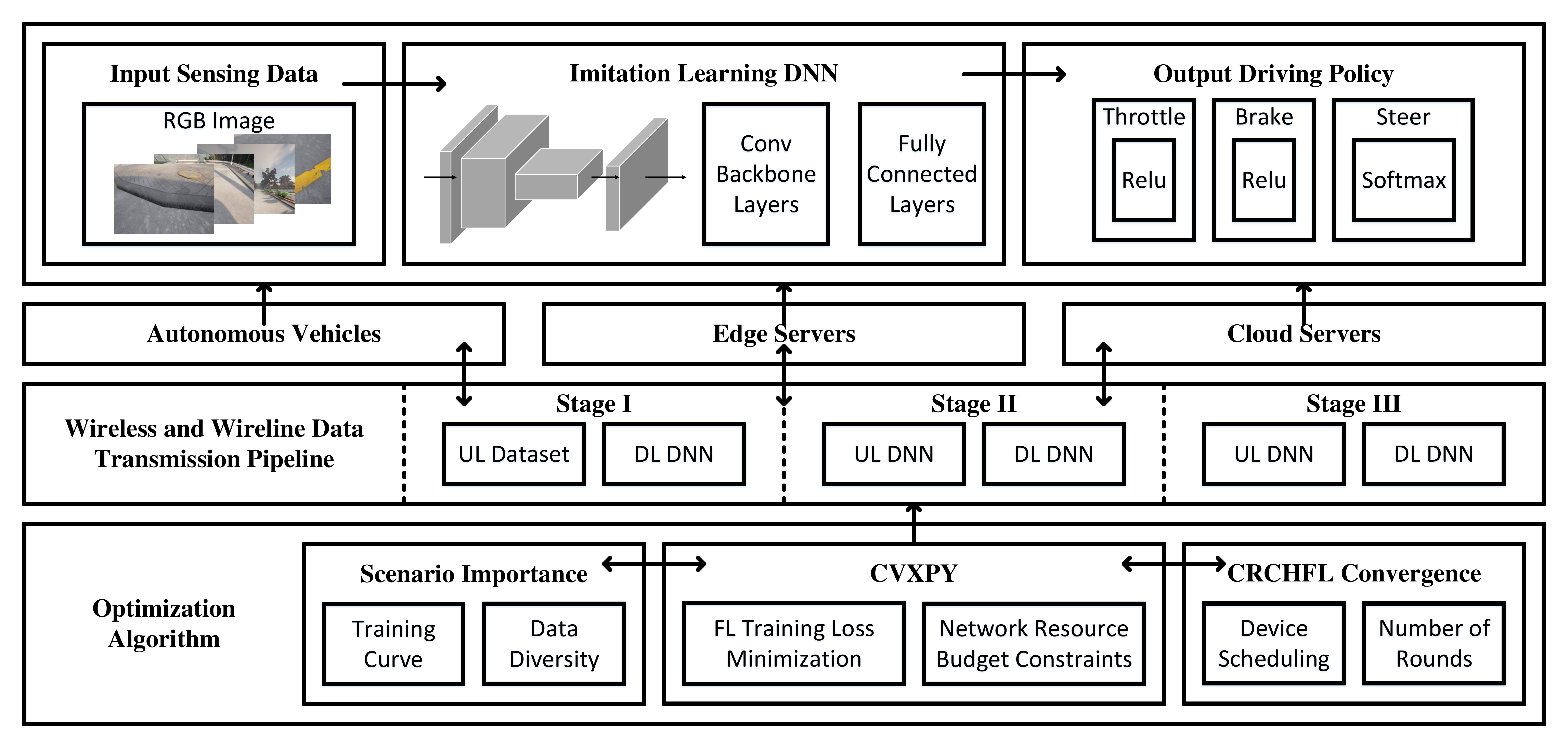}
\vspace{-0.5cm}
\caption{The structure of the proposed CRCHFL. The first layer (top layer) is overview of imitation learning pipeline, including input sensor data, imitation learning DNN model and predicted output driving actions. Second layer showcases the CRCHFL-involved nodes, i.e., autonomous vehicles, edge servers and cloud server. The third layer is about the communication of CRCHFL, including data and model upload (UL) and download (DL). The fourth layer (bottom layer) focuses on optimization algorithm to schedule communication resources.}
\label{Fig.framework}
\vspace{-0.6cm}
\end{figure*}

\section{System Framework}
We consider a cloud-edge-vehicle system shown in Fig.~\ref{Fig.main}.
The proposed CRCHFL framework is shown in Fig.~\ref{Fig.framework}, which consists of 
an end-to-end imitation learning pipeline (i.e., top of Fig.~\ref{Fig.framework}), a three-stage training procedure (i.e., middle of Fig.~\ref{Fig.framework}) and an optimization-based communication resource scheduler (i.e., bottom of Fig.~\ref{Fig.framework}). 
Specifically, for the $k$-th vehicle ($1\leq k\leq K_n$) in the $n$-th town ($1\leq n\leq N$), denoted vehicle $(n, k)$, 
its DNN model 
$f_{n,k}(\cdot|\mathbf{m}_{n,k})$ with parameter vector $\mathbf{m}_{n,k}$ 
is an inference mapping from a frame of sensor data (e.g., images) $\mathbf{s}_{n,k}$ to a driving actions (i.e., throttle, steer and brake) $\mathbf{a}_{n,k}$, i.e., 
$\mathbf{a}_{n,k}=f_{n,k}(\mathbf{s}_{n,k}|\mathbf{m}_{n,k})$.
To optimize model parameters $\mathbf{m}_{n,k}$, we need to define loss function $\mathcal{L}\left(\mathbf{m}_{n,k},\mathbf{s}_{n,k},\mathbf{a}_{n,k}\right)$, and the ego-vehicle training is given by 
\begin{align}
\mathop{\mathrm{min}}_{\substack{\mathbf{m}_{n,k}}}~\mathcal{L}_{n,k}=
\frac{1}{|\mathcal{T}_{n,k}|}
\sum_{
\left(\mathbf{s}_{n,k}^{(i)},\mathbf{a}_{n,k}^{(i)}\right)\in\mathcal{T}_{n,k}
}\mathcal{L}\left(\mathbf{m}_{n,k},\mathbf{s}_{n,k}^{(i)},\mathbf{a}_{n,k}^{(i)}\right) %,
\label{111}
\end{align}
where $\mathcal{T}_{n,k}$ is the training dataset at vehicle $(n,k)$.

Since the generalization of $f_{n,k}$ increases with the size of dataset $\mathcal{T}_{n,k}$, it is necessary to exploit $\mathcal{T}_{n,k}$ for all $(n,k)$.
However, directly aggregating all datasets would lead to high communication costs and data privacy issues.
To this end, a three-stage training procedure is proposed, which can be divided into cloud pretraining (i.e., stage I), edge FL (i.e., stage II), and cloud FL (i.e., stage III). 

In Stage I, vehicles upload data samples to the cloud server. The cloud server uses collected data to train an initial model, which is then released to all the vehicles and edge servers for subsequent FL. The stage I training is given by  
\begin{align}
%\mathrm{Cloud}:
&\mathop{\mathrm{min}}_{\substack{\mathbf{m}_{cloud,P}}}
\sum_{n=1}^{N} \sum_{k=1}^{K_n}
\frac{1}{|\mathcal{T}_{n,k}^{I}|}
\sum_{(\mathbf{s}^{(i)}_{n,k},\mathbf{a}^{(i)}_{n,k})\in\mathcal{T}_{n,k}^{I}}\hspace{-0.2cm}\mathcal{L}\left(\mathbf{m}_{cloud,P},\mathbf{s}^{(i)}_{n,k},\mathbf{a}^{(i)}_{n,k}\right) %, 
\label{FIL1}
\end{align}
where $\mathcal{T}_{n,k}^{I}$ is the uploaded dataset from vehicle $(n, k)$ to the cloud server. 
The number of samples $\sum|\mathcal{T}_{n,k}^{I}|$ 
is denoted as 
$pretrain\_batch$, which depends on the associated communication resources allocated to Stage I. 

In Stage II, each edge server collects models from associated vehicles and aggregates them according to FedAvg\cite{DBLP:journals/corr/McMahanMRA16}, and then delivers the aggregated model to all associated vehicles. The stage II training is given by
\begin{align}
\mathrm{Vehicle}:&\mathop{\mathrm{min}}_{\substack{\mathbf{m}_{n,k}}}~\mathcal{L}_{n,k},~\forall k,n, \label{FIL1}
\end{align}
\vspace{-0.5cm}
\begin{align}
\mathrm{Edge}:&\mathop{\mathrm{min}}_{\substack{\mathbf{m}_{n,1}=\cdots=\mathbf{m}_{n,{K_n}}}} \mathcal{L}_{edge,n}=\sum_{k=1}^{K_n} \frac{|\mathcal{T}_{n,k}|}{\sum_{k=1}^{K_n}|\mathcal{T}_{n,k}|} \mathcal{L}_{n,k}%,~\forall n,k, 
\label{FIL2}
\end{align}
Equations \eqref{FIL1} and \eqref{FIL2} are executed iteratively and each vehicle local training iterations between two adjacent edge aggregations is denoted as $edge\_interval$, which determines the associated communication resources allocated to Stage II. 

In Stage III, cloud server collects models from edge servers to aggregate together also using FedAvg for producing global model. Then the global model is released to all edge servers and vehicles. The stage III training is given by
\begin{align}
%&\mathrm{Cloud}:\nonumber \\
&\mathop{\mathrm{min}}_{\substack{\mathbf{m}_{edge,1}=\cdots=\mathbf{m}_{edge, N}}} \mathcal{L}_{cloud}=\sum_{n=1}^{N} \frac{\sum_{k=1}^{K_n}|\mathcal{T}_{n,k}|}{\sum_{n=1}^{N}\sum_{k=1}^{K_n}|\mathcal{T}_{n,k}|} \mathcal{L}_{edge, n} %,~\forall n, 
\label{FIL4}
\end{align}
The Stages III and II are executed iteratively and the edge aggregation times of each edge server between two adjacent cloud aggregations is denoted as $cloud\_rounds$, which determines the associated communication resources allocated to Stage III. 

Our proposed CRCHFL framework aims to minimize the generalization error by setting four hyper-parameters: $edge\_interval$,  $cloud\_interval$, $pretrain\_batch$ and $cloud\_rounds$ which determines the limited communication resource distribution, are needed to optimize by an optimization algorithm.
The entire procedure of CRCHFL is summairzed in Algorithm ~\ref{alg:HFL}.
The next section will present details of the resource optimization algorithm.

\setlength{\textfloatsep}{0pt}
\begin{algorithm}[h]
\caption{CRCHFL}
\label{alg:HFL}
\SetAlgoLined
\KwIn{$\mathbf{s}_{n, k}$, $\mathbf{a}_{n,k}$, $\mathcal{T}_{n,k}^{I}$, where n = $1, 2, \cdots, {N}$, and k = $1, 2, \cdots, {K_n}$}
\KwOut{cloud model parameters $\mathbf{m}_{cloud}$}
\nl run CVXPY to output $edge\_interval$, $cloud\_interval$, $cloud\_rounds$ and $pretrain\_batch$ \\
\nl initialize cloud model randomly $\mathbf{m}_{cloud}$ = $\mathbf{m}_{rand}$ \\
\nl collect pretraining data \{$\mathcal{T}_{n,k}^{I}$\}   \\
\nl model centralized pretraining to get $\mathbf{m}_{cloud,P}$ \\
\nl model release: $\mathbf{m}_{1,1}=\cdots=\mathbf{m}_{N, K_N}=\mathbf{m}_{edge,1}=\cdots=\mathbf{m}_{edge,N}=\mathbf{m}_{cloud,P}$ \\
\nl \For {round $i=1,\cdots,cloud\_rounds$}{
    \nl \For{Edge\ Server $n=1,\cdots,N$ \textbf{in parallel}}{
        \nl \For {edge\_agg $\tau_2=1,\cdots,cloud\_interval$}{
    	     \nl \For {Vehicle $k=1,\cdots,K_n$ \textbf{in parallel}}{
                      \nl  \For {epoch $\tau_1=1,\cdots,edge\_interval$}{
                         \nl model updates: $\mathbf{m}_{n,k} \leftarrow  \{\mathbf{s}_{n,k}, \mathbf{a}_{n,k}$\} \\
                          \nl \If{$\tau_1$ == $edge\_interval$}{
    	               \nl    $\mathbf{m}_{n,k} \Rightarrow Edge\ Server\ n$\\
                          }
                        }
	            }
                    \nl\If{$\tau_2$ == $cloud\_interval$}{
                    \nl    $\mathbf{m}_{edge,n}=\sum_{k=1}^{K_n}\frac{|\mathcal{T}_{n,k}|}{\sum_{k=1}^{K_n}|\mathcal{T}_{n,k}|}\mathbf{m}_{n,k}$ \\
    	        \nl    $\mathbf{m}_{edge,n} \Rightarrow Cloud\  Server$ \\
                    }
                    \nl\Else {
                    \nl    $\mathbf{m}_{edge,n}=\sum_{k=1}^{K_n}\frac{|\mathcal{T}_{n,k}|}{\sum_{k=1}^{K_n}|\mathcal{T}_{n,k}|}\mathbf{m}_{n,k}$ \\
    	        \nl    $\mathbf{m}_{edge,n} \Rightarrow All\ associated\ Vehicles$ \\         
                    }
            }
        }
    \nl $\mathbf{m}_{cloud}=\sum_{n=1}^{N}\frac{\sum_{k=1}^{K_n}|\mathcal{T}_{n,k}|}{\sum_{n=1}^{N}\sum_{k=1}^{K_n}|\mathcal{T}_{n,k}|}\mathbf{m}_{edge,n}$ \\
    \nl $\mathbf{m}_{cloud} \Rightarrow All\ Edge\ Servers\ \Rightarrow\  All\ Vehicles$ \\ 
}
\end{algorithm}

\section{Resource Optimization for CRCHFL}
\label{HFL}
This section presents how to model the resource optimization algorithm for the proposed CRCHFL. 
This algorithm is designed to schedule communication resources to transfer model parameters $\mathbf{m}_{edge, n}$, $\mathbf{m}_{cloud}$, and $\mathbf{m}_{n, k}$ in stage II and III, as well as pretraining data in stage I by optimizing $edge\_interval$, $cloud\_interval$, $pretrain\_batch$ and $cloud\_rounds$.

Let the vector \textbf{u}$=[u_{1},u_{2}]^T$ represents the communication throughput allocation for data transfer and model transfer.
Vector \textbf{u} satisfies $u_1 + u_2$ $\leq$ $U_{\mathrm{sum}}$, where $U_{\mathrm{sum}}$ (Unit: \textbf{GB}) is the total throughput budget.
The number of samples and federated learning rounds are determined by their associated communication resources and there exists a tradeoff among stages. For data transfer, data samples should be uploaded from all vehicles to the cloud server. The number of samples $x$ is constrained by communication throughput as $x\leq \frac{u_{1}}{D}$, where $D$ (Unit: \textbf{MB}) is the size of each sample. 
For model transfer, the number of cloud aggregation $t$ is constrained by $t\leq \frac{u_2}{2EM}$, where $E$ (Unit: \textbf{MB}) is the size of DNN model, $M$ is the number of model transfer among nodes (including vehicles, edge servers and cloud server) between two adjacent cloud aggregations, number $2$ means that uplink and downlink are both considered. The total number of effective samples $y$ in stages II and III is proportional to $M$.

The generalization loss of DNNs is proved to be a monotonically decreasing function of the number of effective samples \cite{Chen2021AJL}. Effective samples refer to non-repeated data that is in-the-distribution of the target application domain. Therefore, the key is to compute the number of effective samples $S$ in CRCHFL framework. Ideally, $S$ is the summation of the number of samples in all stages, i.e., $S=x+y$. However, due to the practical limitations of pretraining and federated learning, the following discounting factors should be added:
\begin{itemize}
\item[1)] Pretraining in stage I is at the cloud, but inference is at the vehicles. Therefore, a discounting factor $\alpha\in[0,1]$ should be applied to $x$ owing to domain shifting. Here $x$ actually corresponds to $pretrain\_batch$.

\item[2)] Federated learning is iterative procedure and its convergence rate is $\mathcal{O}(\frac{1}{t})$\cite{SEUNGHS1992SOLF}. We need to apply discounting factor $(1- dt^{-1})$ to $y$ for finite iterations, where $d$ is parameter representing the convergence rate of federated learning. For CRCHFL, $t$ is decomposed into three factors $I_e$, $I_c$ and $T$. Here $I_e$ corresponds to $edge\_interval$, $I_c$ corresponds to $cloud\_interval$ and $T$ corresponds to $cloud\_rounds$.

\item[3)] Even if the federated learning converges, there exists a gap between its performance and that of centralized learning due to the distributed datasets. Therefore, discounting factors $\gamma\in[0,1]$ is applied to $y$.
\end{itemize}
Combining the above observations, the joint resource allocation is formulated as
\begin{align}
\mathcal{P}:\mathop{\mathrm{max}}_{\{x,\ I_e,\ I_c,\ T\}}~&
\alpha x + \gamma(1- d(I_eI_cT)^{-1}) y\nonumber
\\
\mathrm{s.t.}~~~&x\leq \frac{u_{1}}{D},\,0 < T \leq \frac{u_2}{2EM}, 0 < I_c < I_e,
\nonumber
\\
&x\geq0,\,y\geq0,\, u_1\geq 0, u_2 > 0,
\nonumber
\\
&u_{1} + u_2 = U_{sum}.
\label{positive} 
\end{align}
Since $S$ is a concave function of $(x, ~I_eI_cT)$ and the constraints are linear, $\mathcal{P}$ can be solved by tree search over ($I_e,I_c$) and off-the-shelf numerical solvers CVXPY\cite{diamond2016cvxpy}. We should note that the outputs of optimization algorithm $x, I_e, I_c, T$ are utilized to set $pretrain\_batch$, $edge\_interval$, $cloud\_interval$ and $cloud\_rounds$, respectively.  

\section{Experimental Results}

This section is divided into four parts: \textbf{A. General setup}. In this part, a detailed setup of experiments is introduced, including data sampling, adopted imitation learning (IL) model structure and so forth.  \textbf{B. Performance comparison of CRCHFL, HFL and SFL}. In this part, we will verify whether the highly elaborated CRCHFL framework can stand out compared to HFL and SFL. \textbf{C. Simulation comparison of CRCHFL, HFL and SFL}. In this part, we will simulate CRCHFL, HFL and SFL on CARLA platform to further verify the effectiveness and robustness of our proposed CRCHFL. \textbf{D. Ablation study of CRCHFL}. In this part, we will compare the performance of different settings of CRCHFL, and also analyze the results.

\subsection{General Setup}
\label{General_setup}

\begin{figure*}[!t]
\vspace{-1.0cm}
\centering 
\includegraphics[width=170mm, height=85mm]{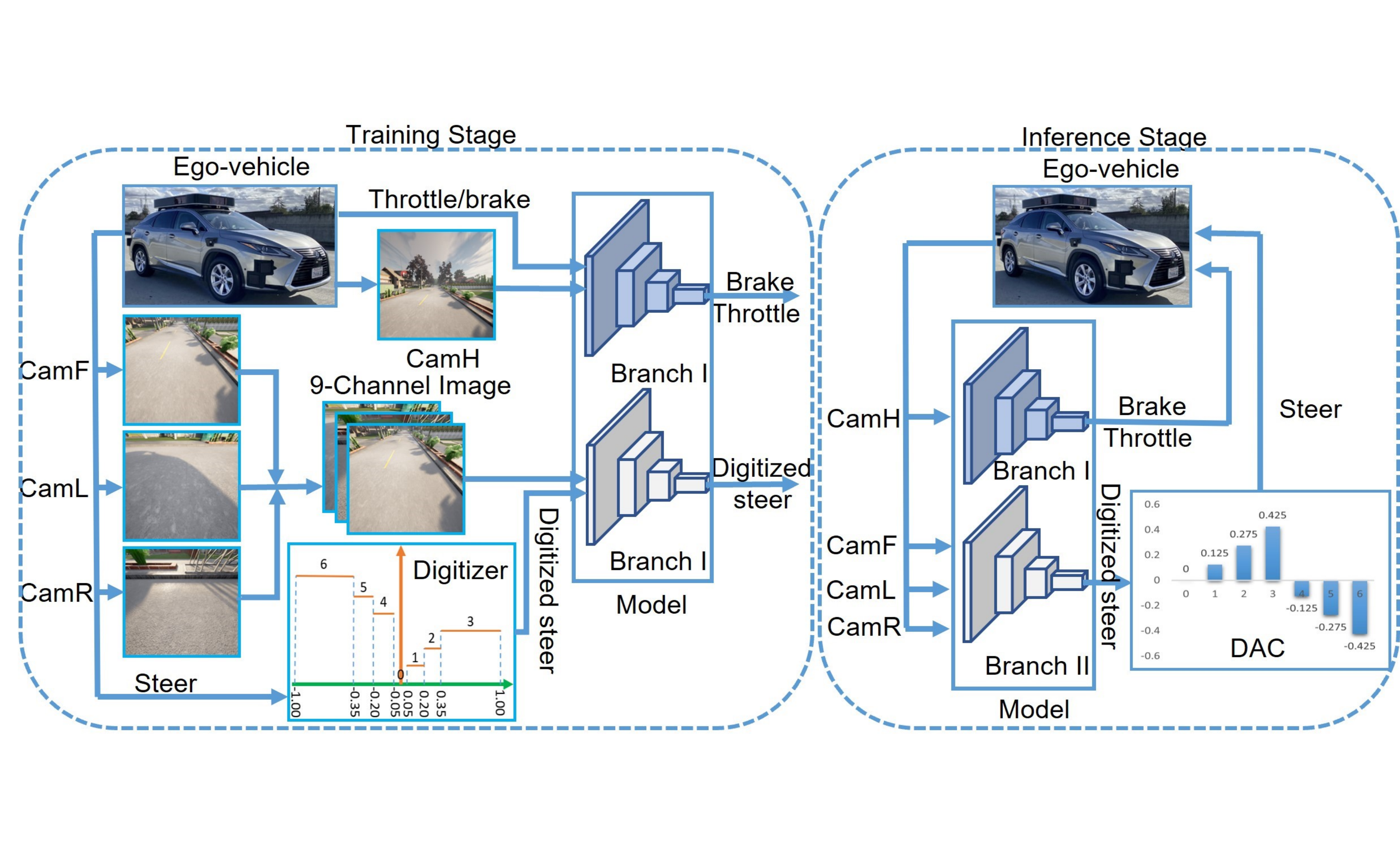}
\vspace{-1.2cm}
\caption{Illustration of entire training and inference process of autonomous driving vehicle. $\mathbf{Digitizer}$ is used to quantize the steer action into 7-level digital signal, while $\mathbf{DAC}$ is utilized to convert the predicted digitized steer to an analogical steer to drive the ego-vehicle. The proposed $\mathbf{Model}$ contains two mutually independent branches where $\mathbf{Branch\ I}$ is responsible for brake and throttle signal and $\mathbf{Branch\ II}$ is used to predict the steer signal.}
\vspace{-0.6cm}
\label{Fig.entire_process}
\end{figure*}

We adopt Town01 and Town02 maps in CARLA to generate training and testing dataset for verification of our proposed CRCHFL scheme. Datasets, both including raw images and actions, are recorded simultaneously when the vehicles driven by CARLA auto-pilot mode are moving forward.
To be specific, 2 vehicles are spawned in different zones of Town01 to record 5856 training samples (i.e., each vehicle contains 2928 training samples) and 3 vehicles spawned in Town02 record 2586, 2587, and 1555 data samples, respectively. The testing dataset contains 2081 samples in total, with 1061 from Town01 and 1020 from Town02.
Each sample consists of a 3D action vector (throttle, steer, brake) produced by expert drivers treated as ground truth and 4 images captured by cameras $\mathbf{CamH}$, $\mathbf{CamF}$, $\mathbf{CamR}$ and $\mathbf{CamL}$.
The relative position of these cameras w.r.t. ego-vehicle is represented by a 6D tuple (X, Y, Z, Pitch, Yaw, Roll) (X/Y/Z Unit: m, Pitch/Yaw/Roll Unit: degree). Four cameras are given by $(1.5, 0, 1.5, 0, 0, 0)$, $(2.5, 0, 1.5, -50, 0, 0)$, $(1.0, 1.2, 0.5, -50, 90, 0)$, and $(1.0, -1.2, 0.5, -50, -90, 0)$, respectively.  
We place one edge server at the center of each town, which is denoted as Edge01 for Town01 and Edge02 for Town02. Two edge servers are connected to cloud server that acts as a global fusion center. 
Vehicles inside the same town are directly connected with each other via the associated edge server, while vehicles in different towns have no direct link, i.e., their model sharing is based on multi-hop communications via the edge servers and cloud server. 

The EAD PyTorch framework by NVIDIA \cite{DBLP:journals/corr/BojarskiTDFFGJM16} is adopted as a reference to propose our imitation learning model. Our implementation slightly differs from that in \cite{DBLP:journals/corr/BojarskiTDFFGJM16}: we adopt two branches of individual neural network for predicting steer and throttle/brake, respectively. 
Each branch calculates its loss, gradient, and back-propagation using Adam optimizer separately.
This is because that the steer task is more complex than the throttle/brake task.
Note that only image from $\mathbf{CamH}$ is fed to $\mathbf{Branch\ I}$ for throttle/brake task, while images from $\mathbf{CamF}$, $\mathbf{CamL}$, and $\mathbf{CamR}$ are combined together to build a $9$-channel image into $\mathbf{Branch\ II}$ for steer task. The entire training and inference process of autonomous driving vehicle is demonstrated as in Fig.~\ref{Fig.entire_process}. The system hardware/software configurations and training configurations are listed as Table.~\ref{Tab:configs} and Table.~\ref{Tab:train}, respectively.

\begin{table}[h]
    \centering
    \vspace{-0.4cm}
    \caption{Hardware/Software Configurations}
    \begin{tabular}{cc}
    \hline
        \textbf{Items} & \textbf{Configurations} \\ \hline
        CPU  & AMD Ryzen 9 3900X 12-Core \\ 
        GPU  & NVIDIA GeForce 3090 $\times$ 2 \\ 
        RAM  & DDR4 32G \\ 
        DL Framework  & PyTorch @ 1.13.0+cu116 \\ 
        GPU Driver  & 470.161.03 \\ 
        CUDA  & 11.4 \\ 
        cuDNN  & 8302 \\ \hline
    \end{tabular}
\label{Tab:configs}
\end{table}
\begin{table}[h]
    \centering
    \vspace{-0.8cm}
    \caption{Training Configurations}
    \begin{tabular}{cc}
    \hline
        \textbf{Items} & \textbf{Configurations} \\ \hline
        \multirow{2}{*}{Loss}  & nn.MSE (\textbf{Branch I})  \\ 
        ~ & nn.CrossEntropyLoss (\textbf{Branch II}) \\
        Optimizer  & Adam \\ 
        Adam Betas  & (0.9, 0.999) \\ 
        Weight Decay  & 3e-3 \\ 
        Batch Size  & 32 \\ 
        Learning Rate  & 1e-4 \\ 
        $cloud\_interval$  & 1 \\ 
        \hline
    \end{tabular}
\label{Tab:train}
\vspace{-0.3cm}
\end{table}

In addition, we adopt two metrics for further quantitative evaluation of CRCHFL hereafter. The first metric is the loss summation of \textbf{Branch I} and \textbf{Branch II} and is called \textbf{Loss}, and the second metric is the accuracy of \textbf{Branch II} and is termed \textbf{Accuracy}.

\subsection{Performance Comparison of CRCHFL, HFL and SFL}
\label{performance_of_CRCHFL}

In this part, we will compare CRCHFL with other two benchmarks under fixed communication throughput budget (i.e., 20G) : 1) \textbf{HFL}\cite{app12020670}, which does not collect data samples for pretraining and has no optimization to schedule constrained communication resources; 2) \textbf{SFL}\cite{DBLP:journals/corr/abs-2110-05754}, which only shares the model parameters; 
3) \textbf{CRCHFL}, which is our proposed scheme. 
For each case, training is stopped when remaining communication throughput is not enough to transfer model parameters once, and then the best performance model from the saved results is then used for testing.

\begin{figure*}[h]
\centering 
\subfigure[\textbf{Accuracy}]{
\label{Fig.e2_acc}
\includegraphics[width=75mm, height=43mm]{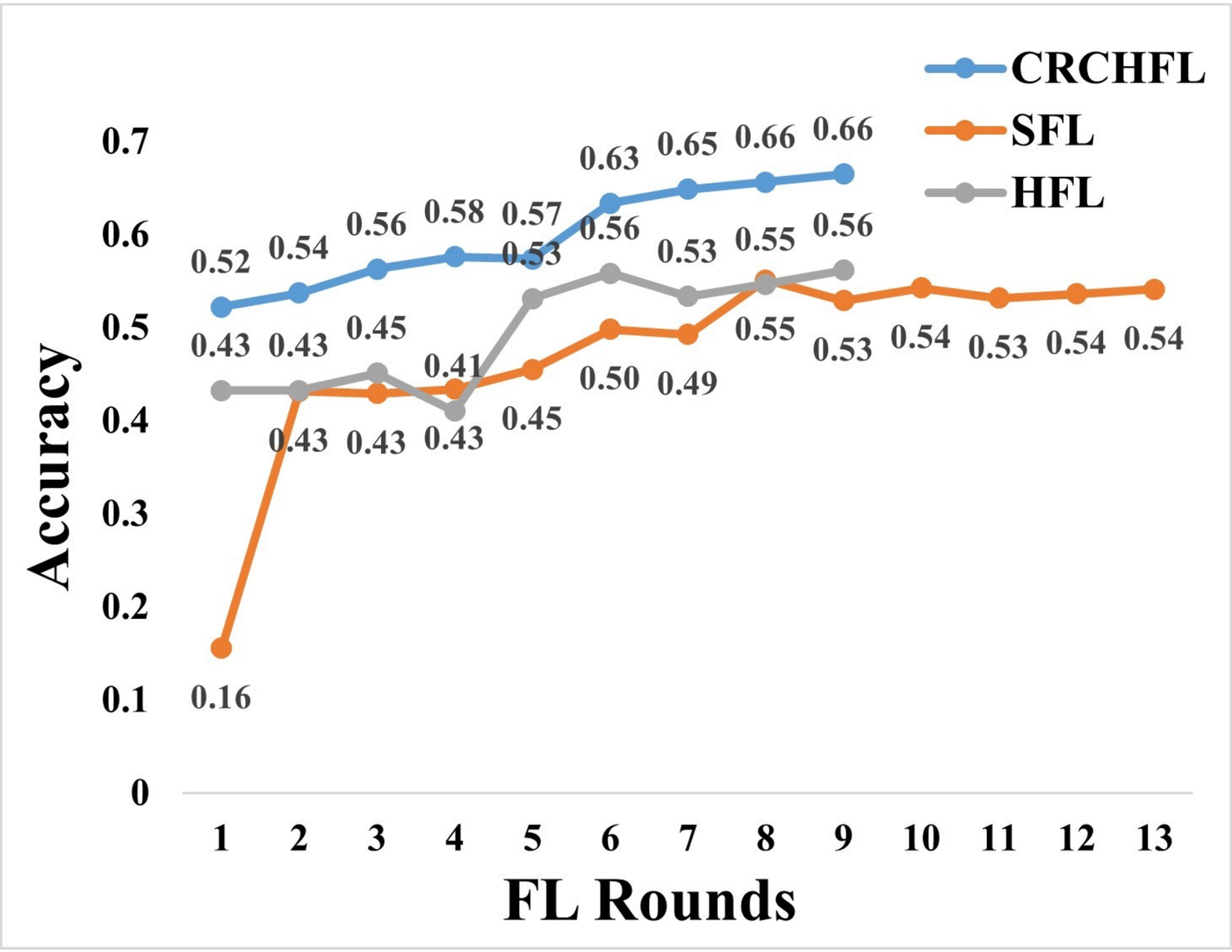}}
%\hspace{-0.8cm}
\subfigure[\textbf{Loss}]{
\label{Fig.e2_loss}
\includegraphics[width=75mm, height=43mm]{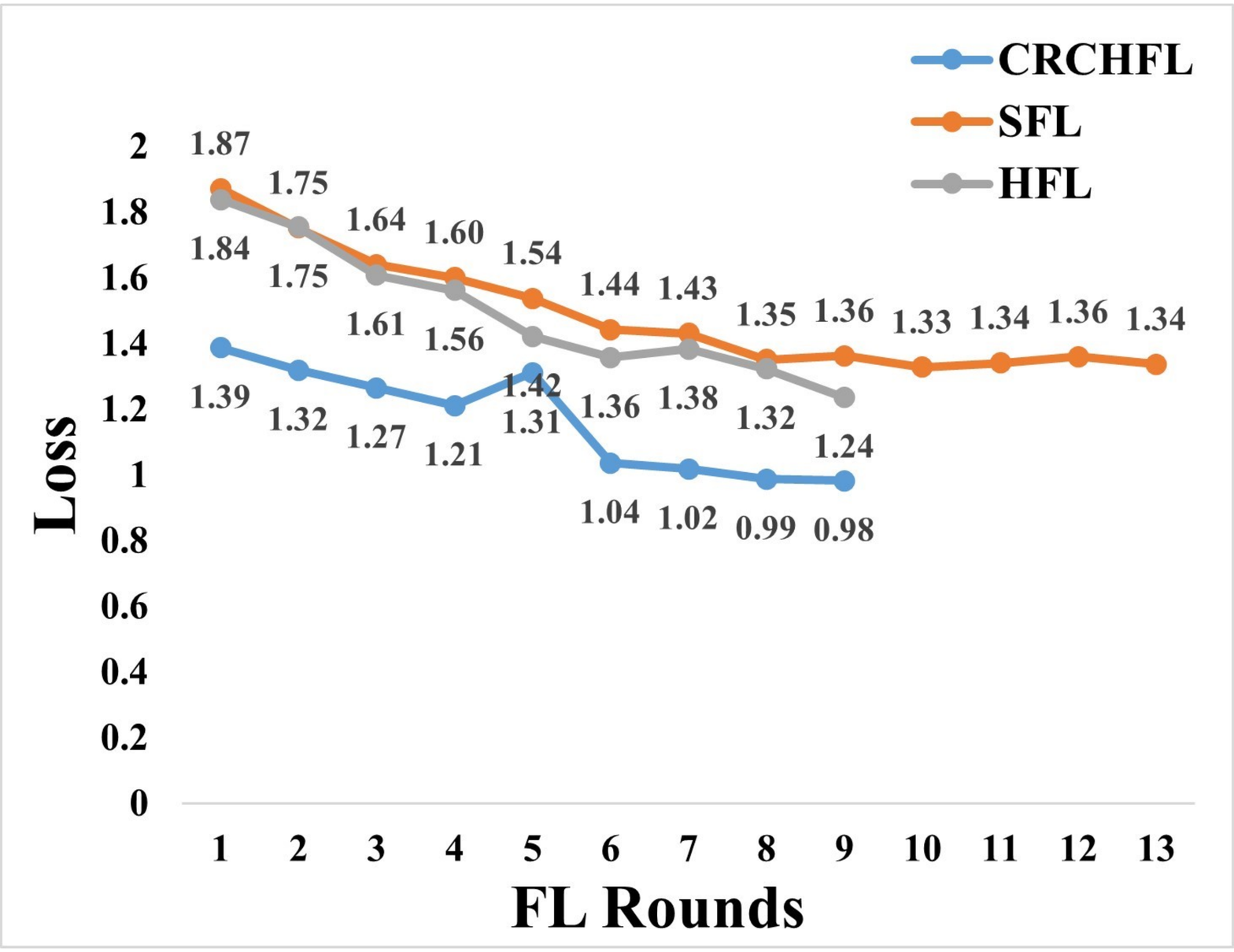}}
\vspace{-0.3cm}
\caption{(a) Comparison of evaluation \textbf{Accuracy} of SFL, HFL and CRCHFL w.r.t FL rounds. (b) Comparison of evaluation \textbf{Loss} of SFL, HFL and CRCHFL w.r.t FL rounds. These experiments are all conducted under 20GB throughput budget.}
\label{metrics1}
\vspace{-0.4cm}
\end{figure*}

\begin{figure*}[h]
\centering 
\subfigure[\textbf{Accuracy}]{
\label{Fig.e2_acc1}
\includegraphics[width=75mm, height=43mm]{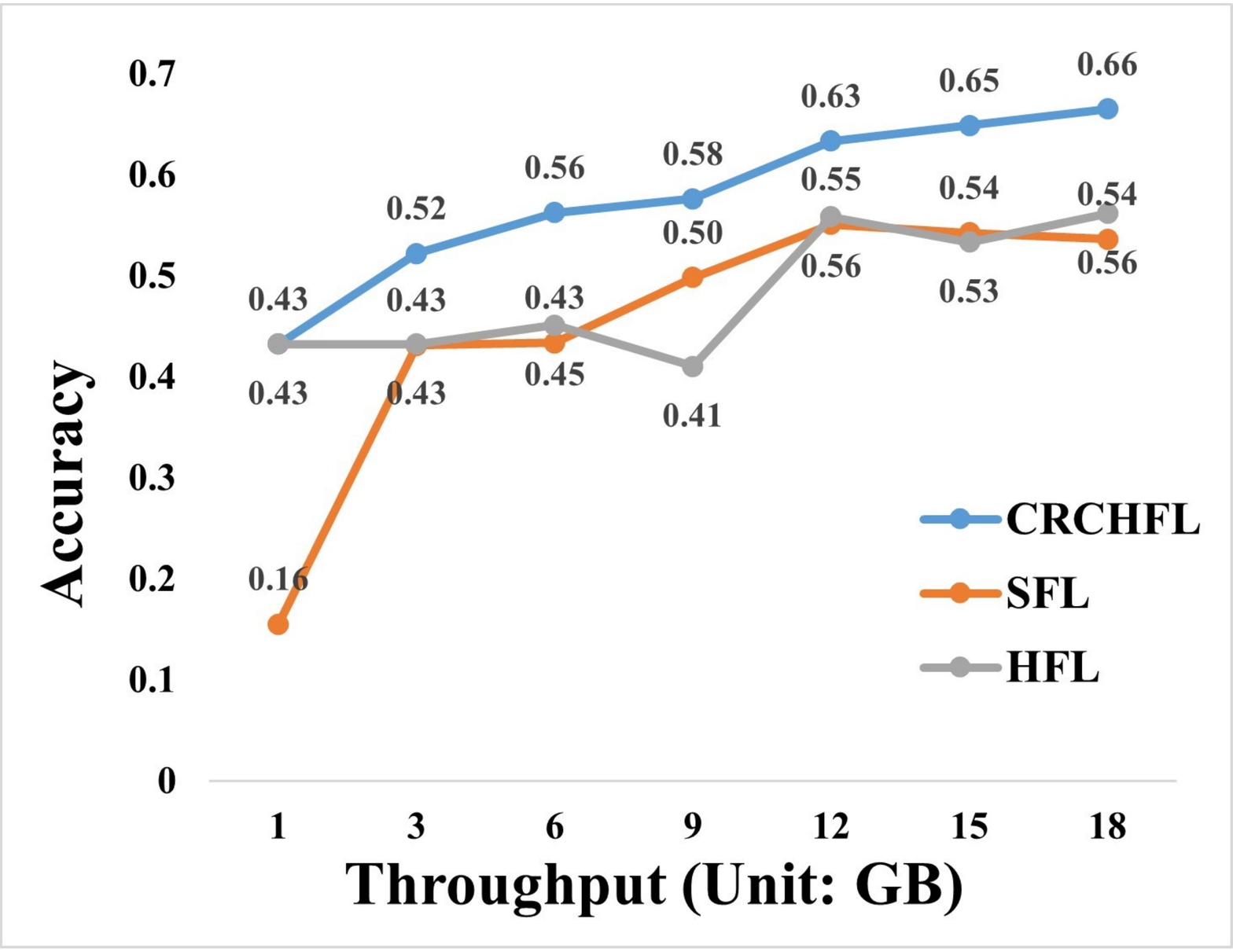}}
%\hspace{-0.8cm}
\subfigure[\textbf{Loss}]{
\label{Fig.e2_loss1}
\includegraphics[width=75mm, height=43mm]{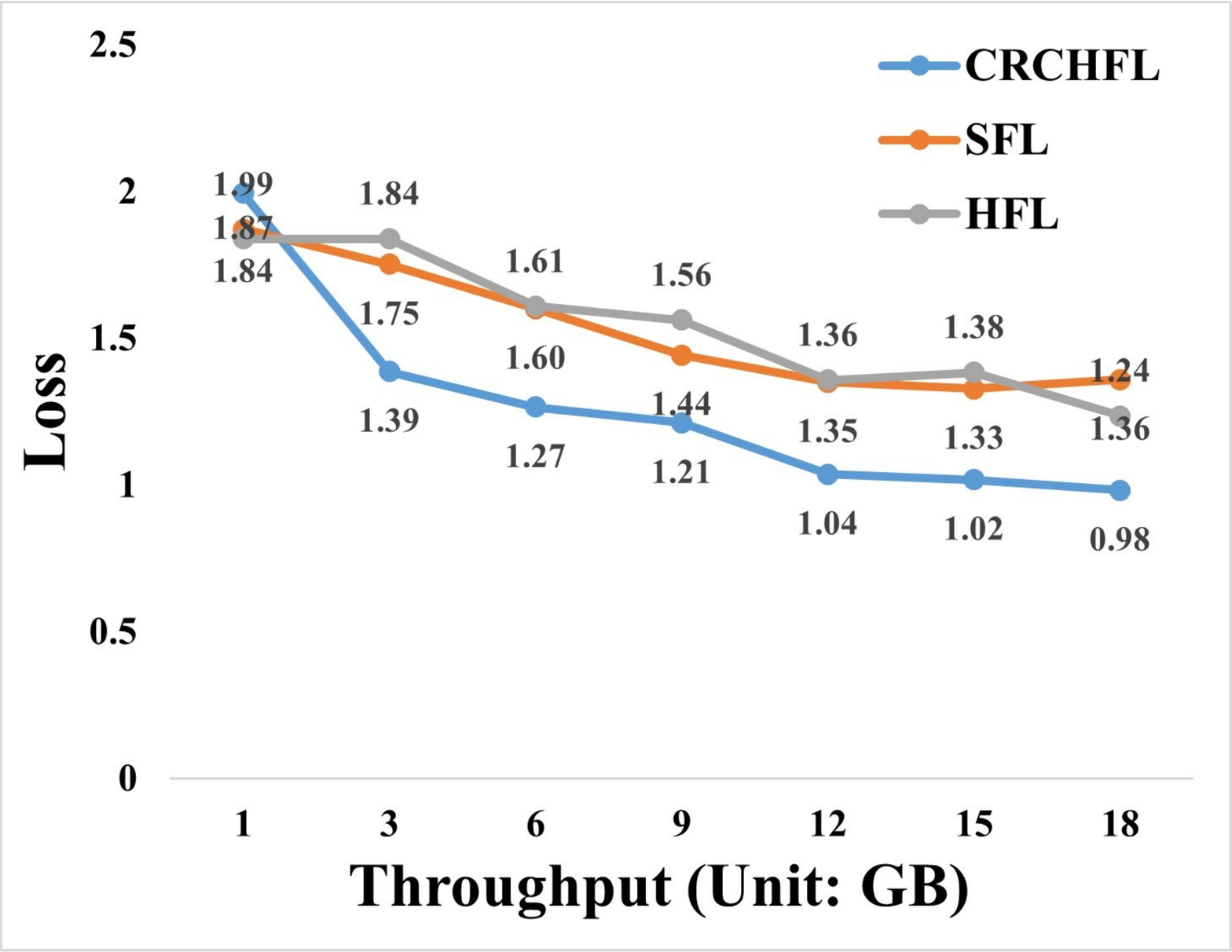}}
\vspace{-0.3cm}
\caption{(a) Comparison of evaluation \textbf{Accuracy} of SFL, HFL and CRCHFL w.r.t consumed throughput in one training process. (b) Comparison of evaluation \textbf{Loss} of SFL, HFL and CRCHFL w.r.t consumed throughput in one training process. These experiments are all conducted under 20GB throughput budget.}
\label{metrics2}
\vspace{-0.6cm}
\end{figure*}

First, Let's compare the performance of CRCHFL, HFL and SFL. With the fixed communication throughput budget (i.e., 20G) and the aforementioned general setup, SFL can perform 13 rounds of cloud aggregation, while HFL can only perform 9 rounds of cloud aggregation because of part of throughput being consumed by communication between edge servers and cloud server. In contrast, although the designed CRCHFL consumes part of the communication throughput when transmitting data samples at the pretraining stage, it can also perform 9 rounds of cloud aggregation due to the proposed optimization algorithm that can schedule communication resources between edge aggregation and cloud aggregation. The inference performance can be found in Fig.~\ref{Fig.e2_acc} and Fig.~\ref{Fig.e2_loss}. It is easy to find that our proposed CRCHFL achieves the best performance in terms of \textbf{Accuracy} and \textbf{Loss}. Specifically, from Fig.~\ref{Fig.e2_acc}, we can find that the \textbf{Accuracy} of CRCHFL is 10.33\% higher than that of HFL and 12.41\% higher than that of SFL, and from Fig.~\ref{Fig.e2_loss}, we can also find that the \textbf{Loss} of CRCHFL is the smallest among them. In addition, due to the introduction of the pretraining stage, it is easy to find that our proposed CRCHFL converges faster than HFL and SFL, so it can be inferred that the designed pretraining stage is necessary and effective.

On the other hand, although the above comparison has illustrated the advantages of our proposed CRCHFL framework, we will further compare these three methods from another perspective where performance is compared when same throughput is consumed by training. Specifically, We further investigate SFL, HFL and CRCHFL by comparing their performance as the consumed throughput increases during one training process. It can be seen from Fig.~\ref{Fig.e2_acc1} and Fig.~\ref{Fig.e2_loss1} that: 1. the performance of all three cases improves as consumed communication throughput increases, which is in line with the actual situation; 2. our proposed scheme not only has better \textbf{Accuracy} and \textbf{Loss} than the other two cases when consuming the same amount of throughput, but also has a more stable evolution trend. This further validates the effectiveness and robustness of our proposed scheme from another perspective.

In summary, just as Fig.~\ref{metrics1} and Fig.~\ref{metrics2}, we evaluate SFL, HFL and CRCHFL from two perspectives. Both results suggest that our elaborated CRCHFL framework has better generalization and faster convergence rate.

\subsection{Simulation Comparison of CRCHFL, HFL and SFL}
\label{simulation_of_CRCHFL}

\begin{figure*}[t]
\vspace{-0.2cm}
\centering  %图片全局居中
\subfigure[CRCHFL @ Town01]{
\label{Fig.fed.1}
\includegraphics[width=0.32\textwidth, height=0.16\textwidth]{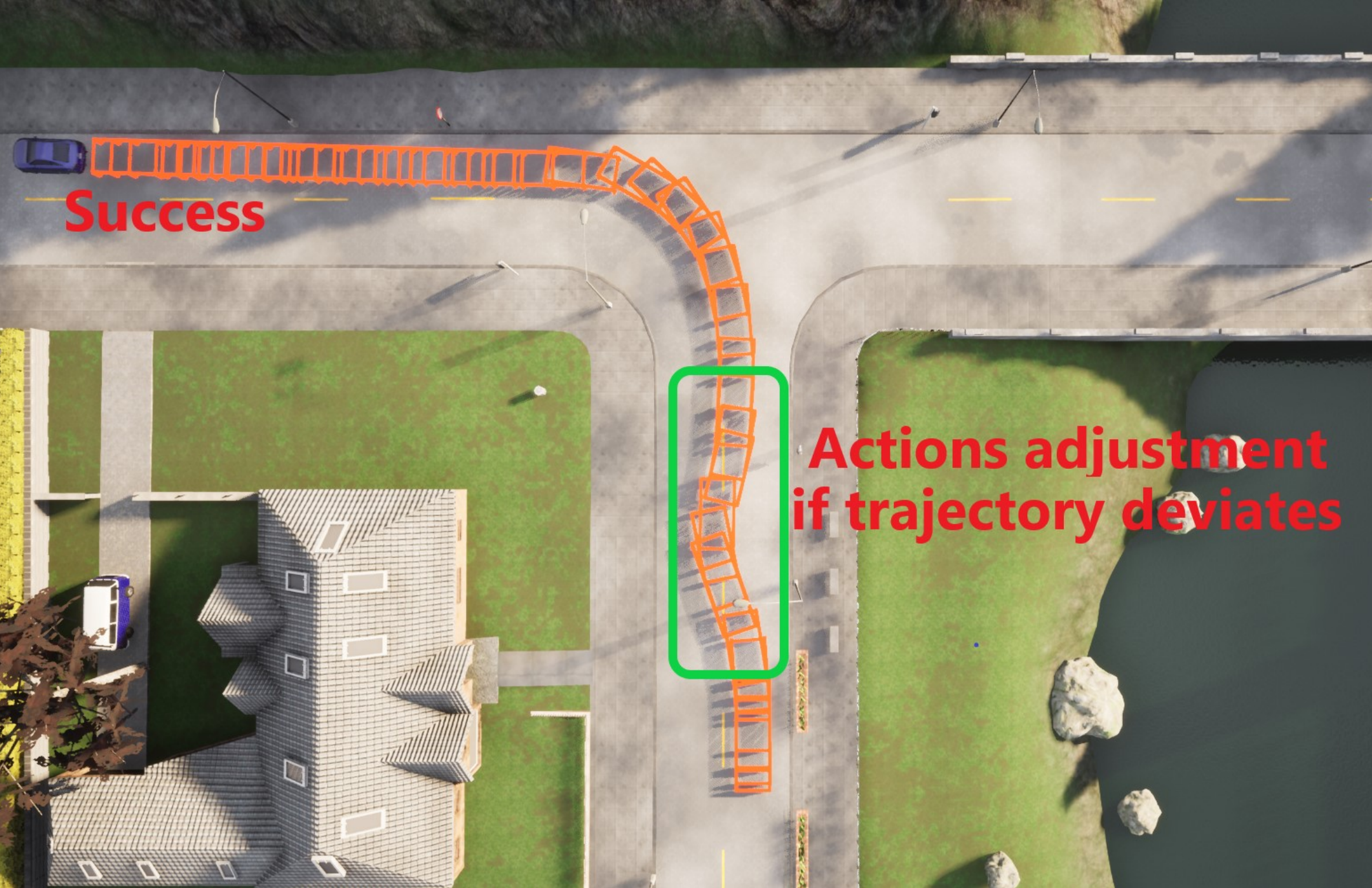}}
\subfigure[HFL @ Town01]{
\label{Fig.fed.2}
\includegraphics[width=0.32\textwidth, height=0.16\textwidth]{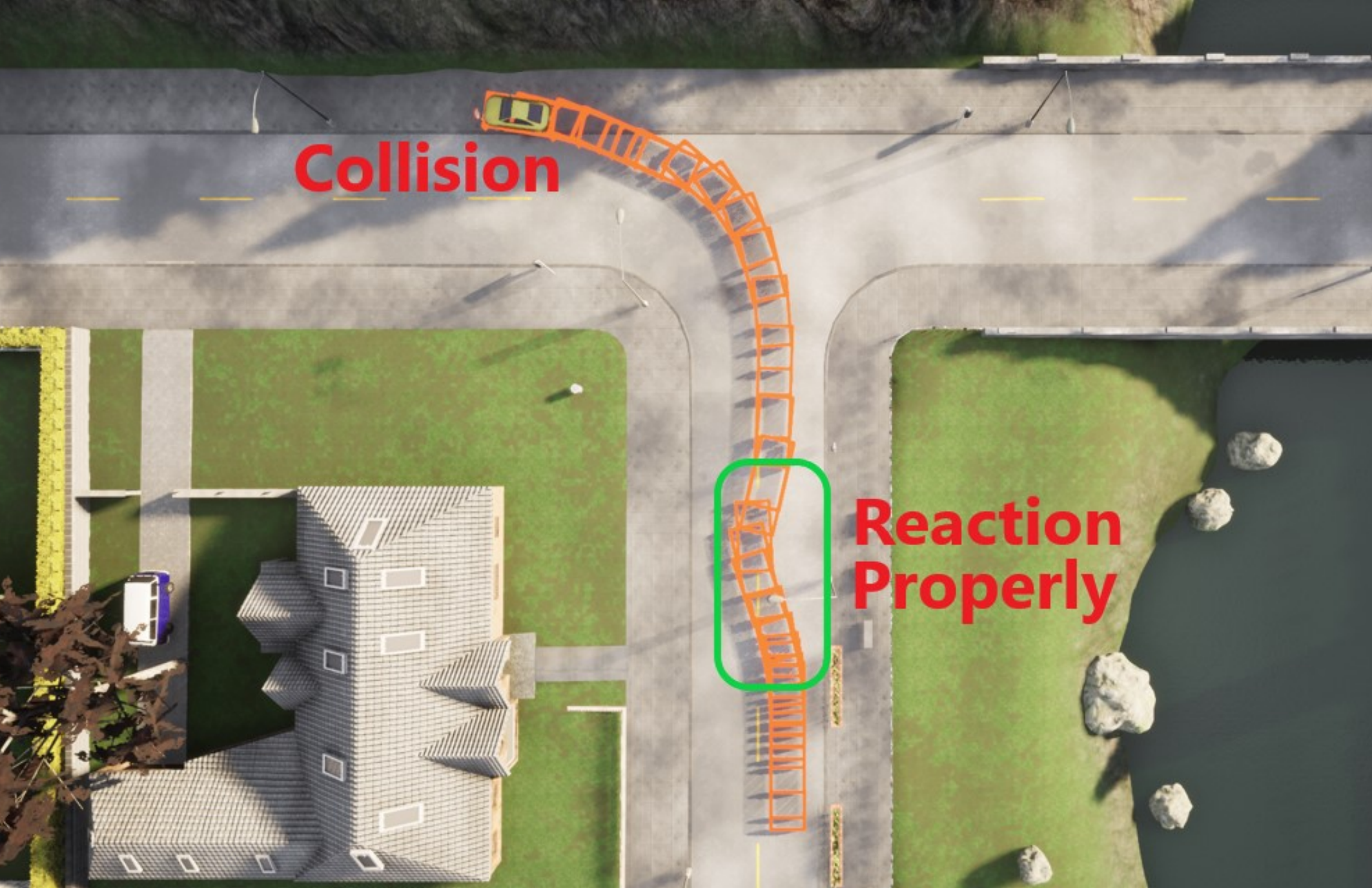}}
\subfigure[SFL @ Town01]{
\label{Fig.fed.3}
\includegraphics[width=0.325\textwidth, height=0.16\textwidth]{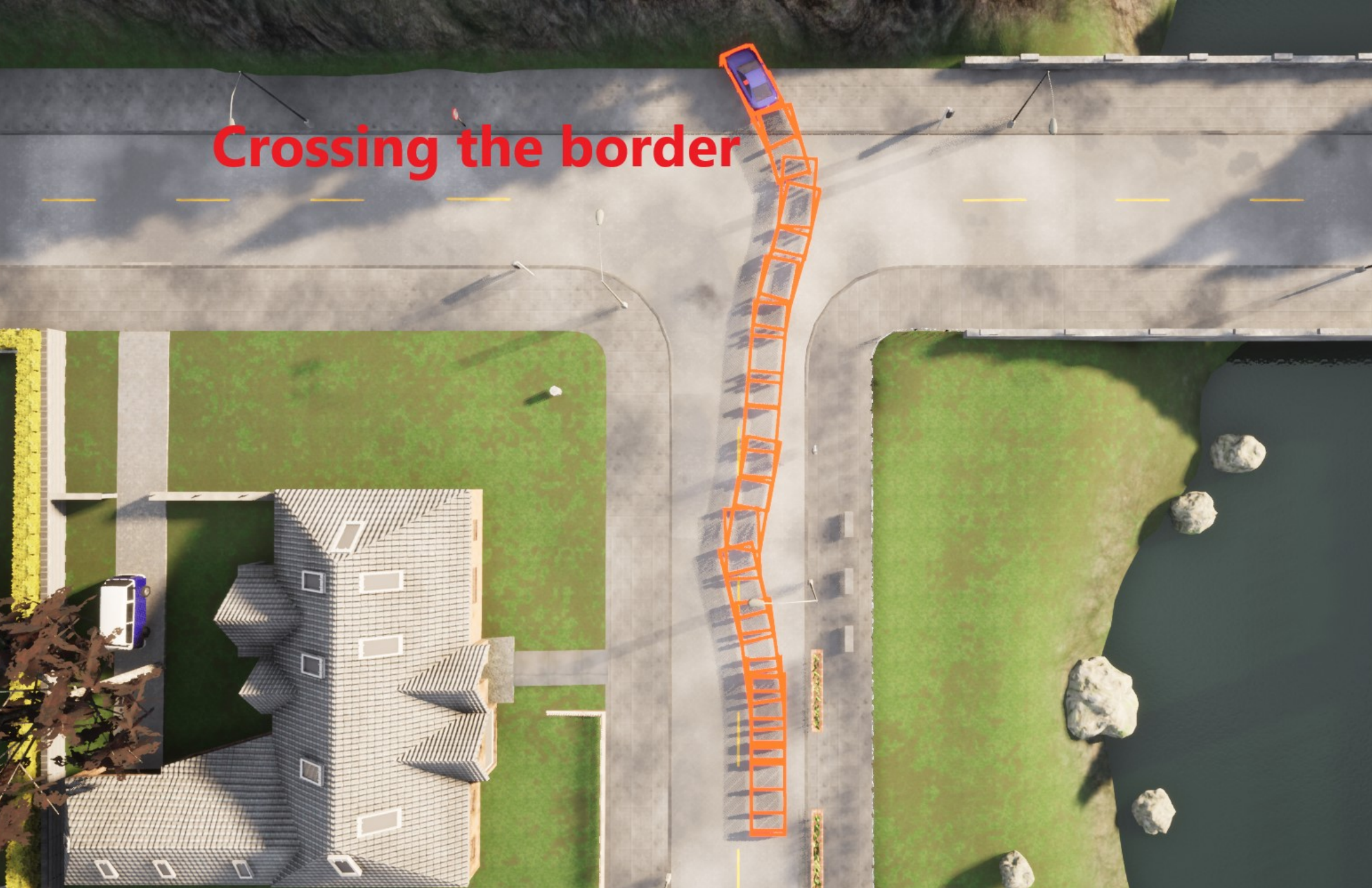}}
\subfigure[CRCHFL @ Town02]{
\label{Fig.fed.4}
\includegraphics[width=0.32\textwidth, height=0.16\textwidth]{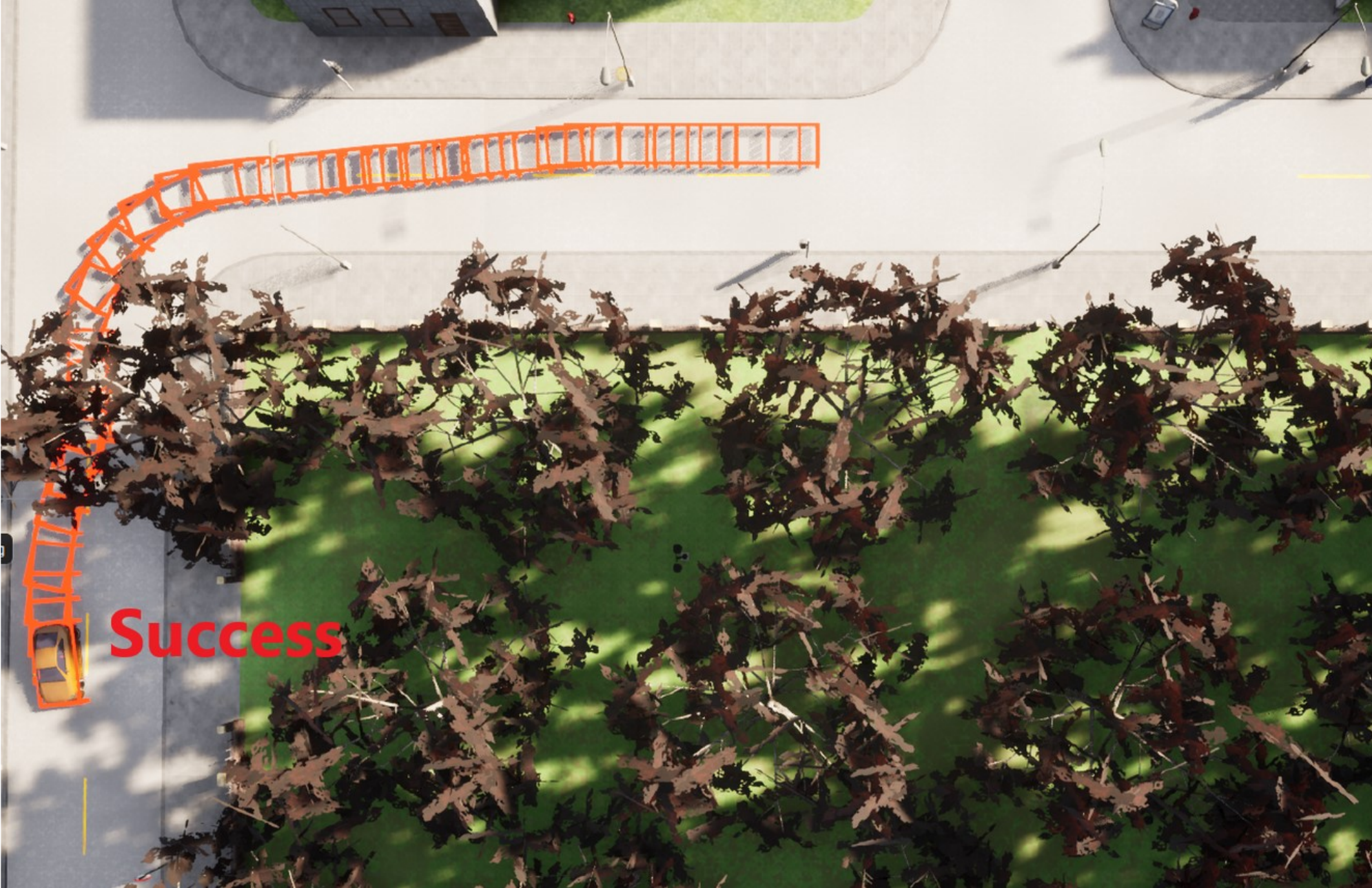}}
\subfigure[HFL @ Town02]{
\label{Fig.fed.5}
\includegraphics[width=0.325\textwidth, height=0.16\textwidth]{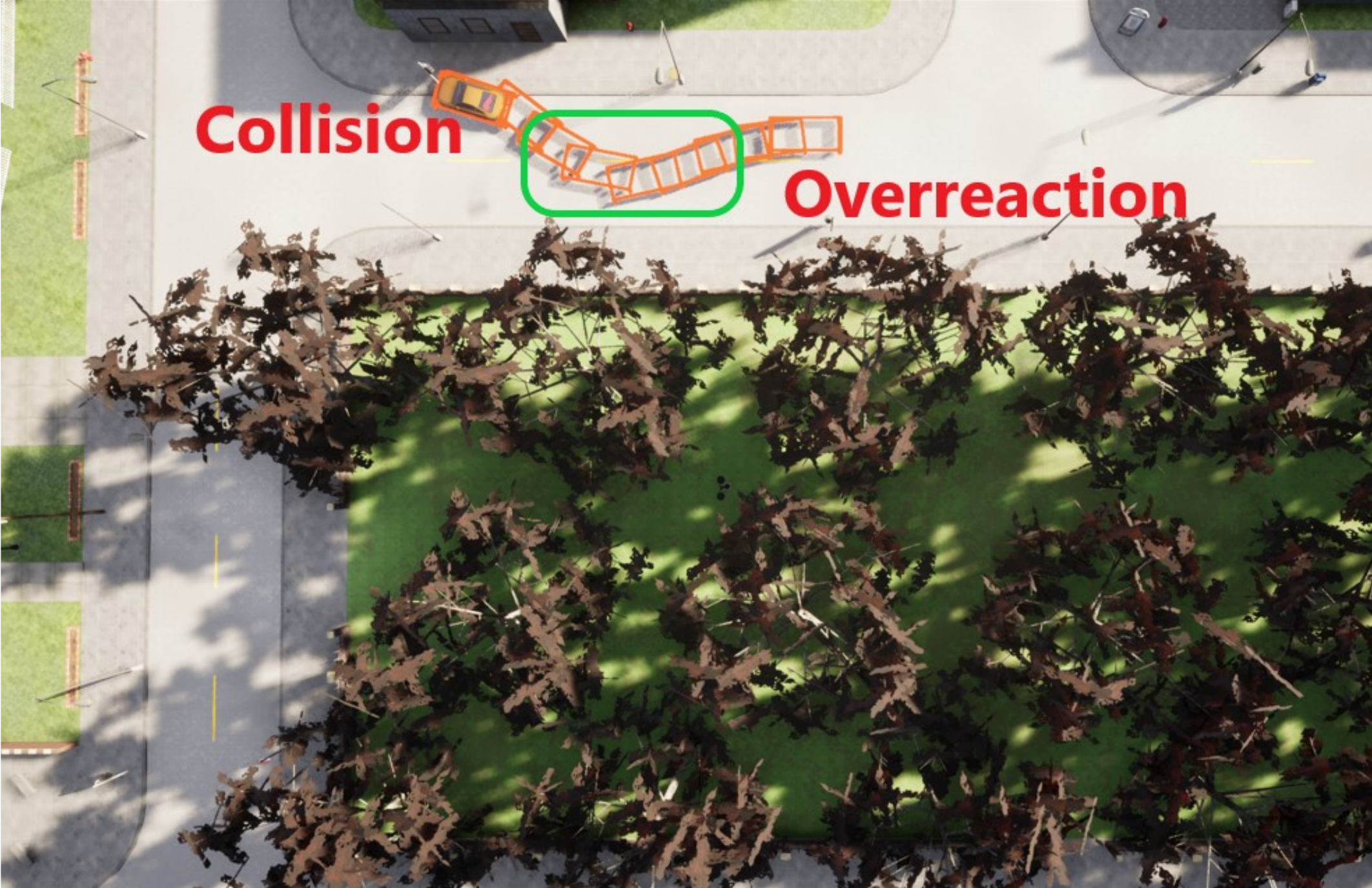}}
\subfigure[SFL @ Town02]{
\label{Fig.fed.6}
\includegraphics[width=0.32\textwidth, height=0.16\textwidth]{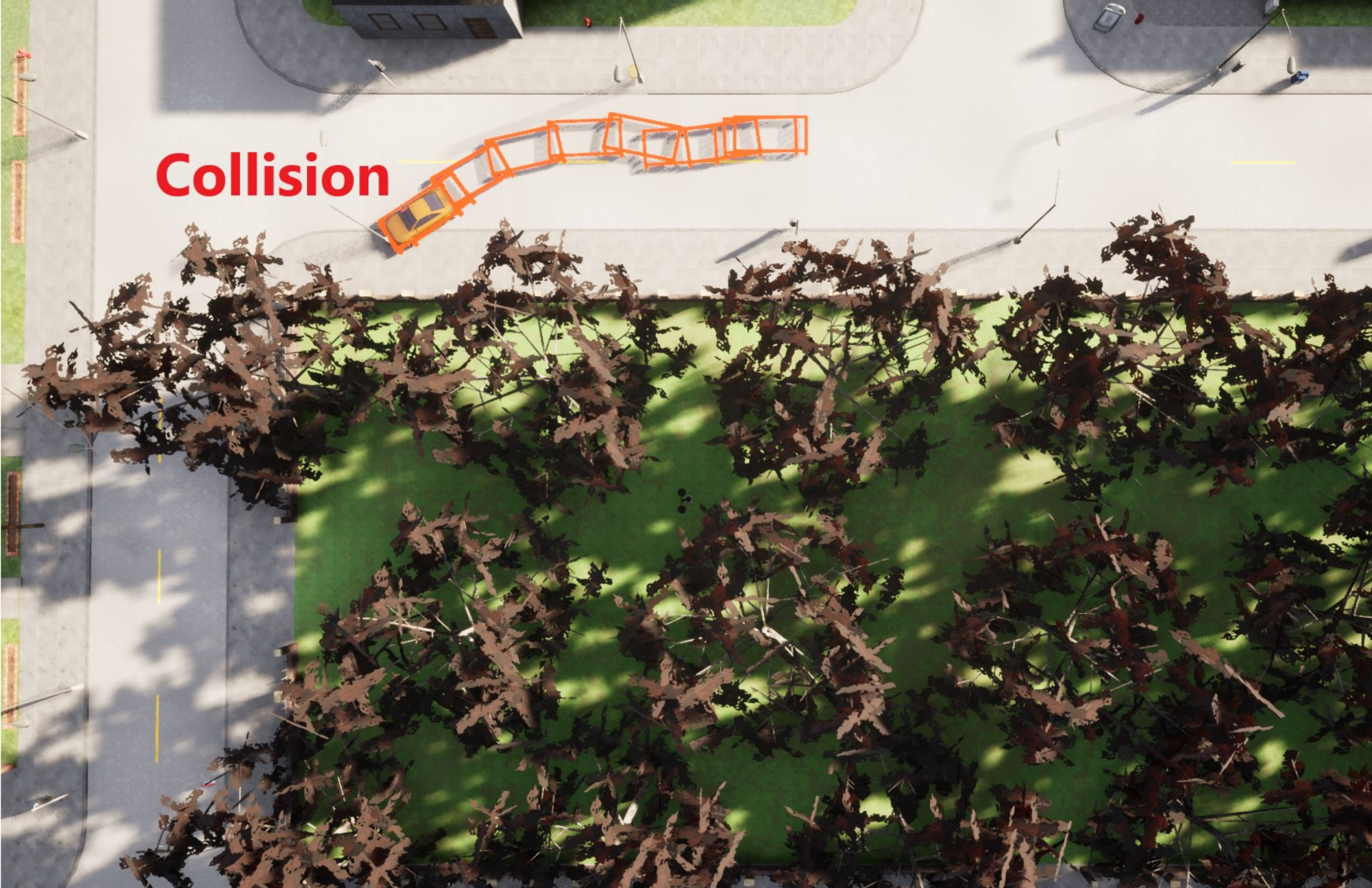}}
\vspace{-0.2cm}
\caption{Simulation Comparison between CRCHFL (a,d), HFL (b,e) and SFL (c,f) on CARLA platform. The boxes in above figures depict the trajectory of ego-vehicle.}
\vspace{-0.5cm}
\label{Fig.sim}
\end{figure*}

\begin{figure*}[h]
\centering  %图片全局居中
\subfigure[Brake]{
\label{Fig.action.1}
\includegraphics[width=0.355\textwidth, height=0.1665\textwidth]{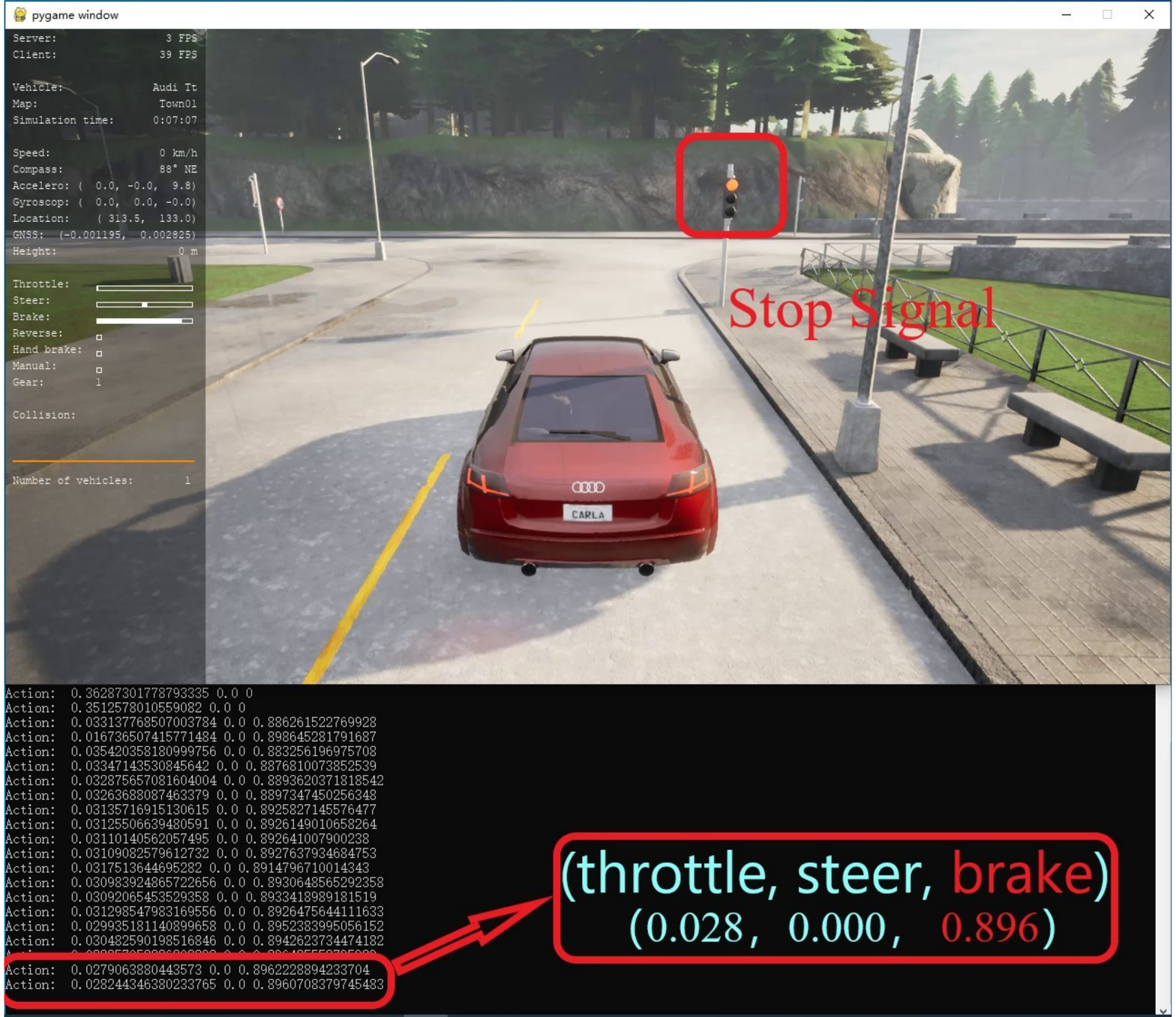}}
\hspace{-1.0cm}
\subfigure[Throttle]{
\label{Fig.action.2}
\includegraphics[width=0.355\textwidth, height=0.1665\textwidth]{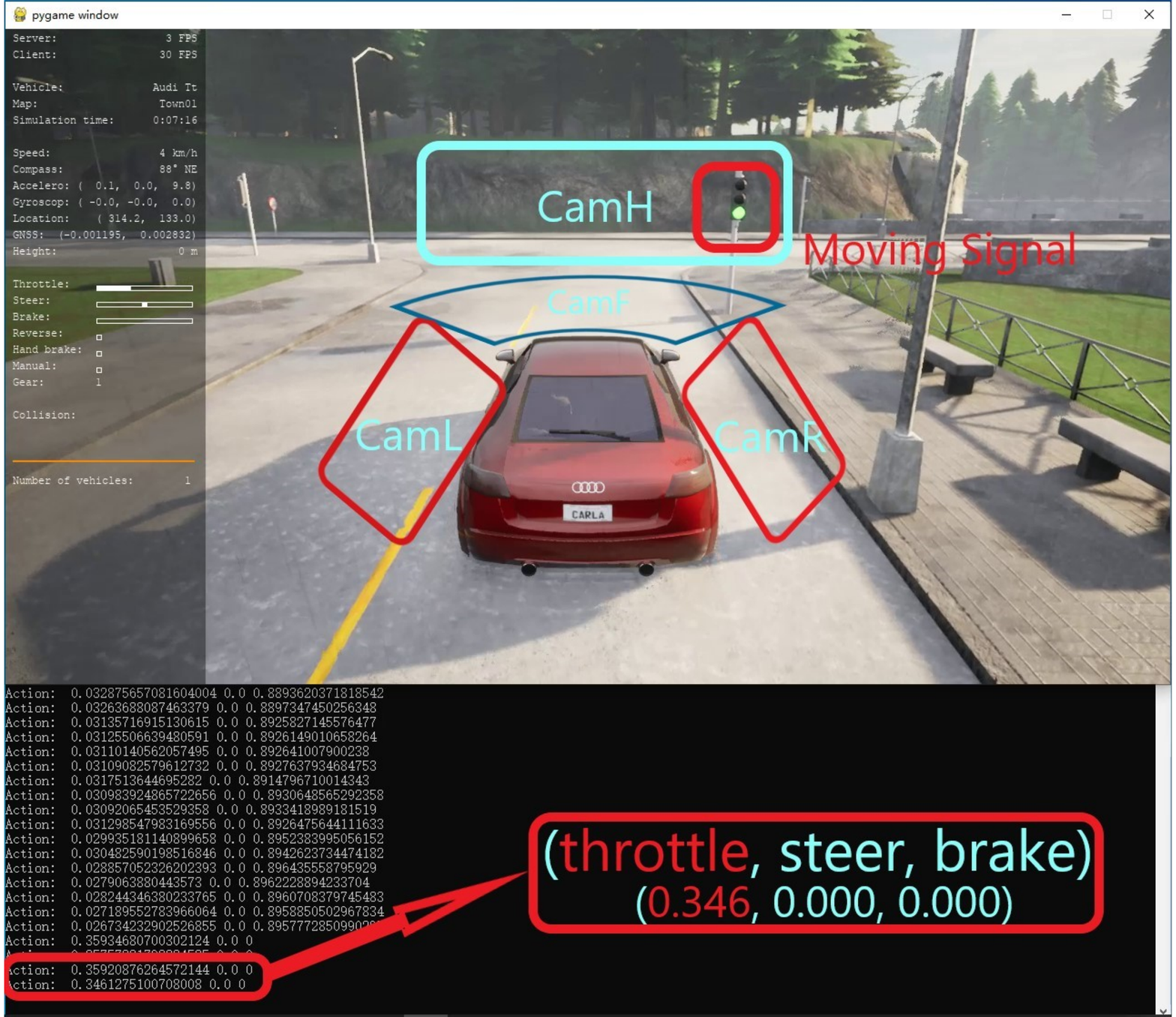}}
\hspace{-1.0cm}
\subfigure[Throttle and Steer]{
\label{Fig.action.3}
\includegraphics[width=0.355\textwidth, height=0.1665\textwidth]{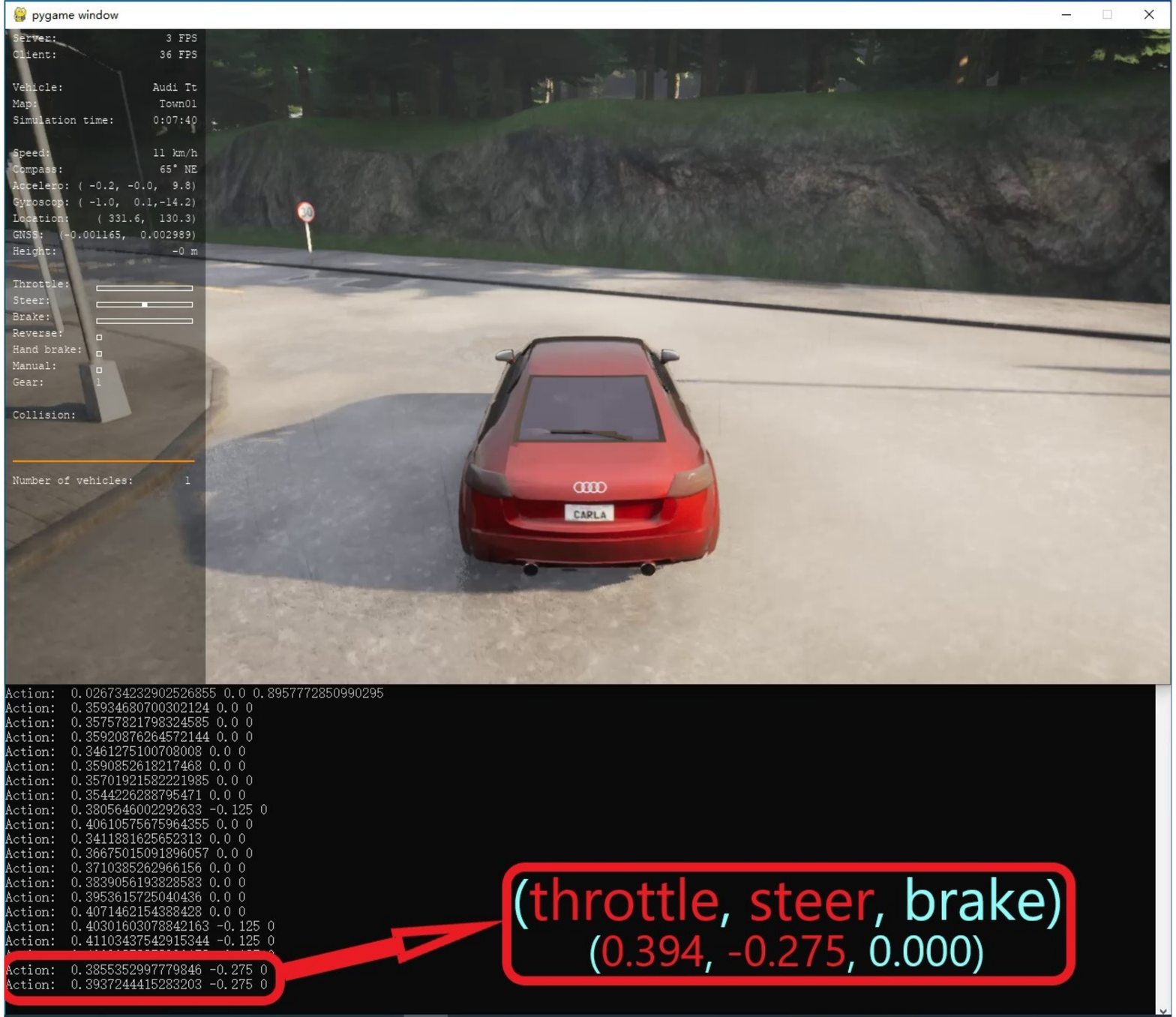}}
\vspace{-0.3cm}
\caption{
(a) Brake action predicted by CRCHFL when the traffic light turns red; (b) Throttle action predicted by CRCHFL when the traffic light turns green;  (c) Throttle-and-steer action predicted by CRCHFL when the vehicle passes through a T-junction. The boxes in the figure (b) illustrate the Field of View (FoV) of four cameras.}
\vspace{-0.5cm}
\label{Fig.action}
\end{figure*}

\begin{figure*}[h]
\centering
\subfigure[Resources Distribution]{
\label{Fig.resource_dist}
\includegraphics[width=0.33\textwidth,height=0.18\textwidth]{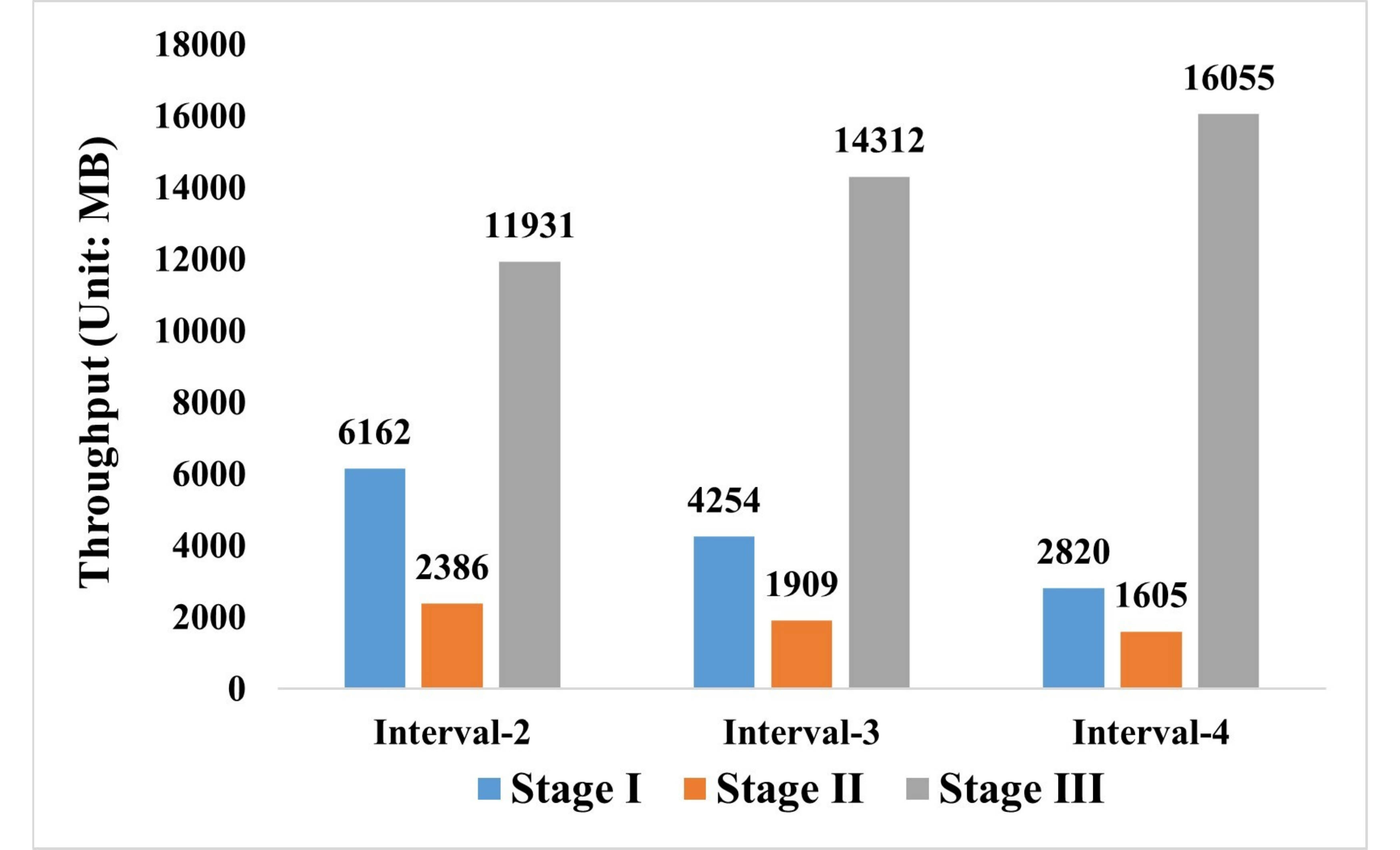}}
\hspace{-0.4cm}
\subfigure[Loss]{
\label{Fig.d_loss}
\includegraphics[width=0.33\textwidth,height=0.18\textwidth]{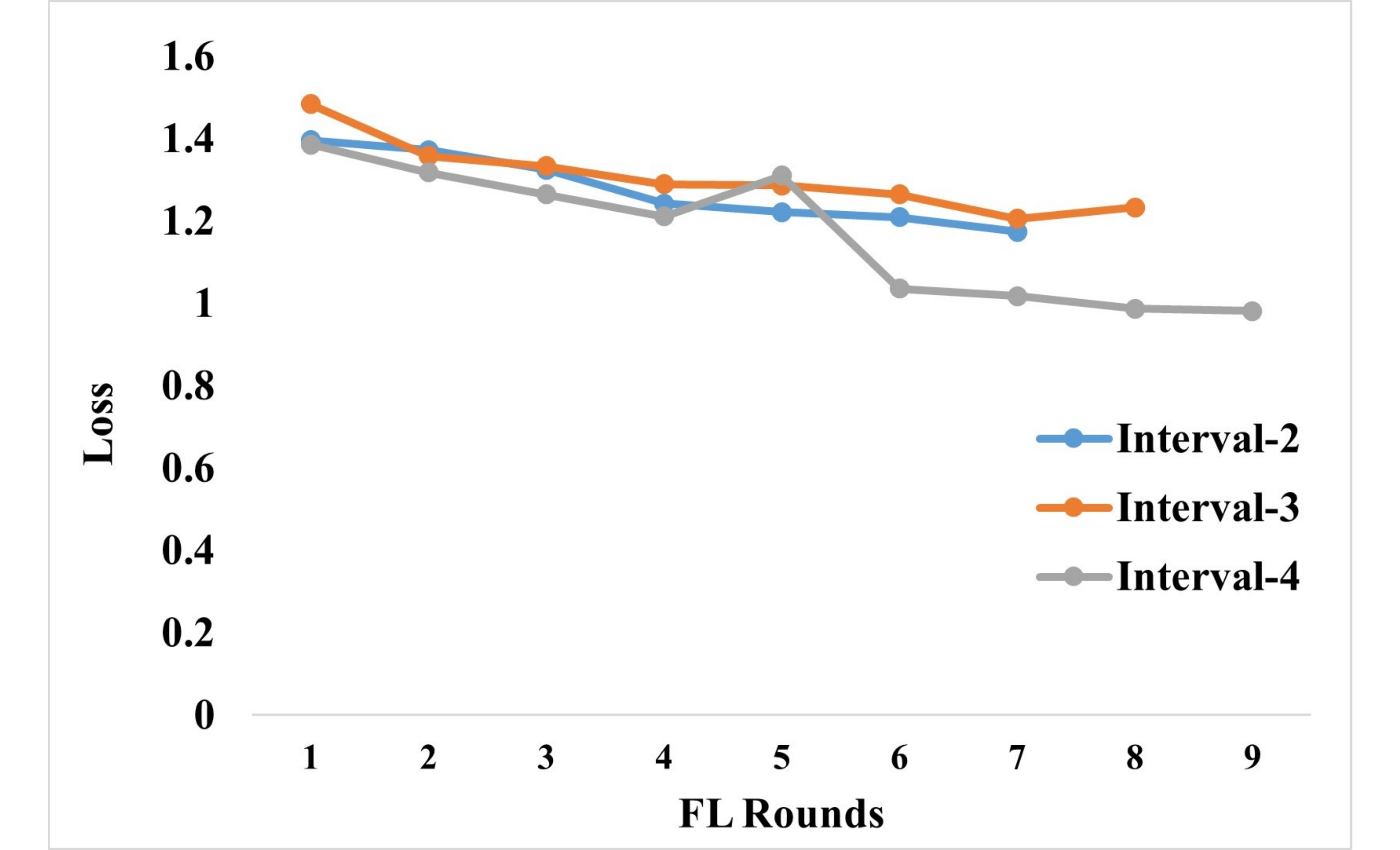}}
\hspace{-0.4cm}
\subfigure[Accuracy]{
\label{Fig.d_acc}
\includegraphics[width=0.33\textwidth,height=0.18\textwidth]{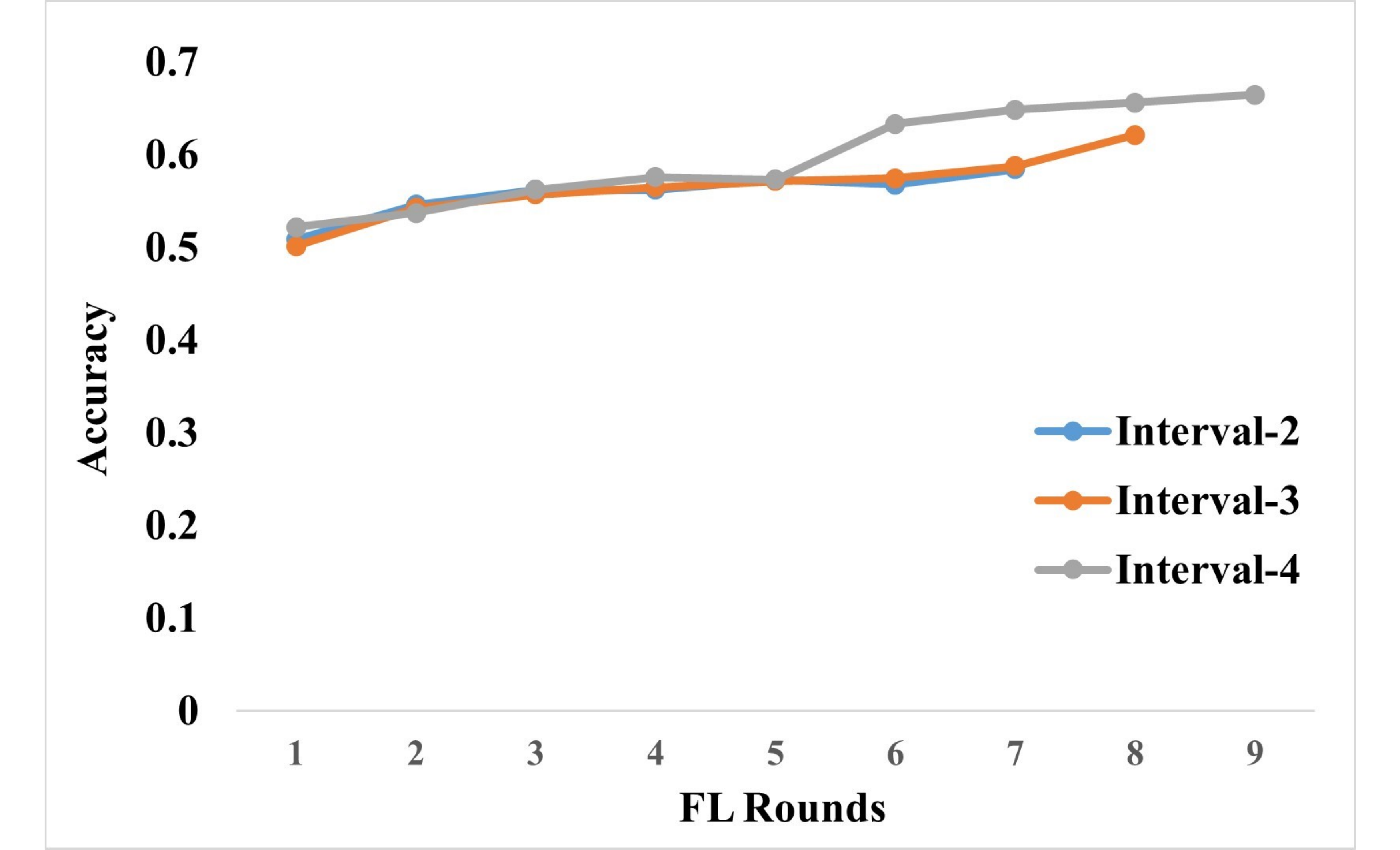}}
\vspace{-0.2cm}
\caption{(a) Illustration of fixed communication throughput budget (i.e., 20G) distribution across stages for different CRCHFL settings. (b) Test loss under different CRCHFL settings. (c) Test accuracy under different CRCHFL settings.}
\vspace{-0.58cm}
\label{Fig.ablation_of_CEHFL}
\end{figure*}

In this part, we simulate the outputs of CRCHFL, HFL and SFL on CARLA platform as well, and the simulation results are shown as Fig.~\ref{Fig.sim}. As it can be seen from Fig.~\ref{Fig.fed.1} and Fig.~\ref{Fig.fed.4}, the CRCHFL-controlled ego-vehicle can pass the T-junction smoothly, and when there is a deviation of the trajectory, the actions can be adjusted in time to return to the correct path. However, as it can be seen from Fig.~\ref{Fig.fed.2} HFL-controlled vehicle can though react properly when the trajectory deviates but collide when pass the T-junction, and from Fig.~\ref{Fig.fed.5} HFL-controlled vehicle collides the pole beside the road owing to overreaction of deviation. From Fig.~\ref{Fig.fed.3} and Fig.~\ref{Fig.fed.6} SFL-controlled vehicle collides or crosses the border while passing through the T-junction. In summary, we can conclude that CRCHFL outperforms HFL and SFL at some scenarios in both Town01 and Town02.

In addition, examples of predicted actions by CRCHFL in inference stage can be checked in Fig.~\ref{Fig.action}. It can be seen from Fig.~\ref{Fig.action.1} and Fig.~\ref{Fig.action.2} that the ego-vehicle stops in front of a red traffic light and accelerates immediately after the light turns green. 
This is because that CRCHFL directly maps the raw data into action vector $(0.028, 0.000, 0.896)$ in Fig.~\ref{Fig.action.1} and $(0.346, 0.000, 0.000)$ in Fig.~\ref{Fig.action.2}, thus realizing the above rule of understanding, and stopping or moving actions. 
In Fig.~\ref{Fig.action.3}, the vehicle successfully turns left at the T-junction by properly combining throttle and steer actions as $(0.393, -0.275, 0.000)$. 

As can be seen from Fig.~\ref{Fig.sim} and Fig.~\ref{Fig.action}, both the macroscopic comparison of the trajectories of SFL, HFL and CRCHFL and the microscopic insight of predicted actions are illustrations of the good performance of our proposed scheme. On top of \ref{performance_of_CRCHFL}, it further proves the feasibility and effectiveness of our proposed method.

\subsection{Ablation Study of CRCHFL}
\label{ablation_of_CRCHFL}
In experiment \ref{performance_of_CRCHFL} and \ref{simulation_of_CRCHFL}, we have verified that the proposed CRCHFL is feasible and effective compared with SFL and HFL in terms of performance and simulation.
In this part, we will do some ablation experiments in different CRCHFL settings. To investigate how the communication resource distribution changes across stages, we can change the conditions of the optimization algorithm to lead to different optimization results. By comparing these outcomes, we can then derive the variation pattern of the communication resource distribution. To be specific, in our experiments, by fixing $edge\_interval$ to $2, 3\ and\ 4$ respectively, the communication resources allocated to each stage change and the distribution is shown in Fig.~\ref{Fig.resource_dist}.

From Fig.~\ref{Fig.resource_dist}, we can find that the communication resource distribution reveals the pattern of change. Specifically, as $edge\_interval$ increases, communication resources are allocated more to Stage III whereas less to Stage I and II, which means that more communication resources are used for cloud aggregation and less communication resources are used for pretraining samples transfer and edge aggregation. 

In order to explore what happened to the performance of CRCHFL when communication resource distribution varies, we can check the \textbf{Loss} and \textbf{Accuracy} of such CRCHFL settings in Fig.~\ref{Fig.d_loss} and Fig.~\ref{Fig.d_acc}. It is obvious that Interval-4 has the highest \textbf{Accuracy} and the lowest \textbf{Loss} because it has the largest FL rounds. This means that when more communication resources are used for cloud aggregation, it helps to improve the performance of the model, which is also in line with our expectation. It is also found that the \textbf{Accuracy} and \textbf{Loss} of the three cases do not differ too much in the first round of federated learning, especially for \textbf{Accuracy}, which also indicates that only a small number of samples are needed to upload in the pretraining stage to accelerate convergence rate.

In conclusion, through the above experiments and simulations, our proposed framework performs significantly better than HFL and SFL with limited communication resources, mainly because: 1. a small number of data samples are uploaded from each vehicle to the cloud server in the pre-training stage to form a centralized dataset which is used to pre-train the model and accelerate its convergence rate; 2. by using optimization-based method to set $edge\_interval$ and $cloud\_interval$, communication resources can be scheduled for more rounds of cloud aggregation, which in turn improves the model performance overall.

\section{Conclusion}

This paper has proposed a communication resources constrained hierarchical federated learning framework, which aims at finding a trade-off between learning performance and communication resources. Experimental results of our proposed framework demonstrated that our proposed CRCHFL algorithm outperforms existing benchmarks especially when the network budget is tight. However, our framework and experiments still have limitations, such as no consideration of communication channel fading, small number of towns and vehicles, etc. Future directions include multi-modal CRCHFL and ROS implementation of CRCHFL. 

%\bibliographystyle{IEEEtran}
%bibliography{root.bib}

\end{document}